\documentclass{article}

\usepackage[numbers]{natbib}
\usepackage[preprint]{neurips_ai4d3_2023}

\usepackage[utf8]{inputenc} 
\usepackage[T1]{fontenc}    
\usepackage{hyperref}       
\usepackage{url}            
\usepackage{booktabs}       
\usepackage{amsfonts}       
\usepackage{nicefrac}       
\usepackage{microtype}      
\usepackage[table,xcdraw]{xcolor}
\usepackage{graphicx}       
\usepackage{arydshln}
\usepackage{upgreek}
\usepackage{fontawesome}
\usepackage{amsmath}
\usepackage{float}
\usepackage{footnote}
\usepackage{rotating}
\usepackage{multirow}
\usepackage{booktabs}

\begin{document}

\title{Towards a more inductive world for drug repurposing approaches}

%

\author{
  Jesus de la Fuente$^{1,2,\dag}$, Guillermo Serrano$^{1,2,\dag}$, Uxía Veleiro$^{1,\dag}$, Mikel Casals$^{2}$ \\ \textbf{Laura Vera$^{1}$}, \textbf{Marija Pizurica$^{3,4}$}, \textbf{Antonio Pineda-Lucena$^{1}$}, \textbf{Idoia Ochoa$^{2,5}$} \\ \textbf{Silve Vicent$^{1}$}, \textbf{Olivier Gevaert$^{3,*}$} and \textbf{Mikel Hernaez$^{1,5,*}$} \\\\
  $^{1}$CIMA University of Navarra, IdiSNA, Pamplona, Spain. \\$^{2}$ TECNUN, University of Navarra, San Sebastián, Spain. \\
  $^{3}$ Stanford Center for Biomedical Informatics Research, Stanford University, California. \\ $^{4}$
  Internet technology and Data science Lab (IDLab), Ghent University, Belgium. \\ $^{5}$ Instituto de Ciencia de los Datos e Inteligencia Artificial (DATAI), University of Navarra, Spain. \\
  $^{*}$ Corresponding authors, $^{\dag}$ These authors have contributed equally.\\\\
  \texttt{jdlfuentec@gmail.com} \quad \texttt{uxveleiro@gmail.com} \quad \texttt{gserranos@unav.es} \\
  \texttt{ogevaert@stanford.edu} \quad \texttt{mhernaez@unav.es}
}
\maketitle

\begin{abstract}
Drug-target interaction (DTI) prediction is a challenging, albeit essential task in drug repurposing. Learning on graph models have drawn special attention as they can significantly reduce drug repurposing costs and time commitment. However, many current approaches require high-demanding additional information besides DTIs that complicates their evaluation process and usability. Additionally, structural differences in the learning architecture of current models hinder their fair benchmarking. In this work, we first perform an in-depth evaluation of current DTI datasets and prediction models through a robust benchmarking process, and show that DTI prediction methods based on transductive models lack generalization and lead to inflated performance when evaluated as previously done in the literature, hence not being suited for drug repurposing approaches. We then propose a novel biologically-driven strategy for negative edge subsampling and show through \emph{in vitro} validation that newly discovered interactions are indeed true. We envision this work as the underpinning for future fair benchmarking and robust model design. All generated resources and tools are publicly available as a python package.
\end{abstract}

\section{Introduction}

Drug discovery aims at finding the most effective pharmacological compound that can target a specific disease-causing mechanism while yielding minimal side effects. Traditionally, predicting drug-target interactions (DTIs) has relied on determining physical parameters between both components, such as the dissociation constant or the inhibitory concentration \citep{patching2014,shuker1996}. However, experimental screenings of compounds have a limited success rate and require time, effort, and elevated monetary costs \citep{dimasi2016}, which considerably hinders the process of finding new drugs interacting with the intended targets.
High throughput sequencing technologies have unveiled thousands of interesting targets with many potential modulators, making the experimental screening of compounds a daunting challenge. 

This new paradigm, fueled by the current availability of large amounts of biological data, has promoted breakthrough deep learning on graphs~\cite{bronstein2021geometric} approaches that have accelerated the first stages of drug discovery pipelines by narrowing down the most promising DTIs \citep{zheng2013,lavecchia2016,olayan2018ddr,peng2021eegdti,zhao2021hyperatt}. 
However, there are currently four critical challenges that prevent proper performance evaluations of newly proposed DTI prediction models: \textbf{i)} Current gold standard datasets for evaluating DTI prediction models, such as the well-known Yamanishi dataset \citep{yamanishi2008prediction} are small, outdated and missing many interactions. \textbf{ii)} State-of-the-art DTI prediction methodologies require additional information to predict novel DTIs, which is generally not readily available and thus restricts their usage and evaluation. \textbf{iii)} Current methods can be divided into inductive or transductive, based on whether they can learn underlying patterns in the data to make predictions on unseen samples (inductive), or whether they directly build a prediction model for the seen ones (transductive). These structural differences make it challenging to fairly compare these methodologies. \textbf{iv)} Current techniques for dealing with the existing positive/negative edge imbalance when training DTI prediction models do not incorporate biological information, potentially affecting subsequent experimental validation. 

To address these limitations, in this work we perform an in-depth evaluation of current state-of-the-art DTI prediction methodologies, taking into account drug repurposing datasets, transductive and inductive learning and DTIs network splitting and subsampling techniques. We demonstrate that designing DTI prediction methods using transductive-based approaches are not optimal, and recommend utilizing inductive-based ones instead. Specifically, we show that a baseline transductive classifier achieves near optimal performance only due to data leakage. Additionally, we introduce a technique based on Root Mean Square Deviation (RMSD) for subsampling negative edges during the construction of the DTI dataset, and show that it can lead to the discovery of true interactions (validated \emph{in vitro}) otherwise missed. Finally, we provide data and tools to ease the design and the fair evaluation of novel DTI methods as we envision this work as the first step towards a community-driven unified benchmark to assess novel DTI prediction approaches. 


\section{Pearls and pitfalls of current DTI datasets and prediction models}\label{sec:review}

\subsection{Overview of current DTI datasets}

In recent years multiple datasets \citep{wishart2006drugbank, zitnik2018biosnap, davis2011comprehensive, liu2007bindingdb} have been generated for in-silico validation of DTI prediction models. While these datasets typically contain a set of targets and their interacting drugs, they strongly differ in the origin of the data, as well as the topology and size of the network (Appx. Table~\ref{tab:statistics}, Appx. Note~\ref{sec:methods:datasets}). For example, the current gold standard datasets were defined by Yamanishi et al. in 2008, which consist of four precise, albeit small datasets (less than 100 edges) divided by protein families: enzymes (E), ion channels (IC), G-protein-coupled receptors (GPCR) and nuclear receptors (NR) \citep{yamanishi2008prediction}. This contrasts with recent datasets derived from DrugBank, such as DrugBank-DTI \citep{wishart2006drugbank} and BIOSNAP \citep{zitnik2018biosnap} which contain more than 15000 edges. In addition, data from drug-target binding affinity experiments has also been used for DTI prediction tasks (e.g., DAVIS \citep{davis2011comprehensive} and BindingDB \citep{liu2007bindingdb} datasets), with the caveat that data must be previously binarized at an arbitrary threshold of affinity.

\textbf{Chemically-driven evaluation of current DTI datasets.} To enable an accurate DTI prediction and avoid introducing a bias towards certain chemical drug categories, datasets should encompass drugs with high chemical diversity and high promiscuity (i.e., high capability to interact with multiple protein families) \cite{dunn2022,mencher2005}. When assessing the aforementioned datasets for these properties, we found that drugs within datasets are indeed chemically diverse (i.e., their pair-wise Tanimoto distance followed a 0-skewed distribution, Appx. Fig.~\ref{suppfig:s2_rmsd_suppl}, \ref{suppfig:s2_tani_suppl}), and promiscuous (Appx. Fig.~\ref{suppfig:s5_sankey_yamanishi}, \ref{suppfig:s5_sankey_dbbd}). Further, datasets should comprise diverse protein families to enable generalization capabilities of DTI models. However, the included protein families are highly variant across datasets, with some containing a wide range (e.g., DrugBank) and others being family-specific (e.g., Yamanishi Enzymes, Appx. Fig.~\ref{suppfig:s3_drugs_yamanishi_heatmap}, \ref{suppfig:s3_drugs_dbbd_heatmap}, \ref{suppfig:s4_proteins_yamanishi_heatmap} and \ref{suppfig:s4_proteins_dbbd_heatmap}). This analysis revealed that while the latest DTI datasets such as DrugBank are suitable for training DTI prediction models, the still-considered gold standard such as Yamanishi should be used with caution since it may introduce bias towards certain protein families.

\subsection{Overview of current DTI prediction models} 

Depending on their learning process on the DTI graph, current DTI models can be classified into two groups: inductive and transductive. Inductive graph learning involves using a set of labeled nodes/edges to learn the underlying data structure, aiming to make predictions on unseen samples leveraging the knowledge acquired during training in the form of weights. Transductive graph learning, on the other hand, does not build a predictive model from seen nodes/edges, as there are no weights that can be used to predict a set of unseen samples. Instead, it uses every sample in the dataset to generate the desired prediction. The following models have been recently shown to achieve state-of-the-art performance \cite{wen2017deep, ozturk2018deepdta}: DTINet \citep{luo2017dtinet}, DDR \citep{olayan2018ddr}, DTiGEMS+ \citep{thafar2020dtigems} and DTi2Vec \citep{thafar2021dti2vec} which fall under the transductive category, and NeoDTI \citep{wan2018neodti}, Moltrans \citep{huang2020moltrans}, Hyper-Attention-DTI \citep{zhao2021hyperatt} and EEG-DTI \citep{peng2021eegdti} which fall into the inductive category (Appx. Table \ref{tab:model-comparison}, Appx. Note \ref{sec:methods:models}). 

\begin{figure}[ht]
\begin{center}
\includegraphics[width=\linewidth]{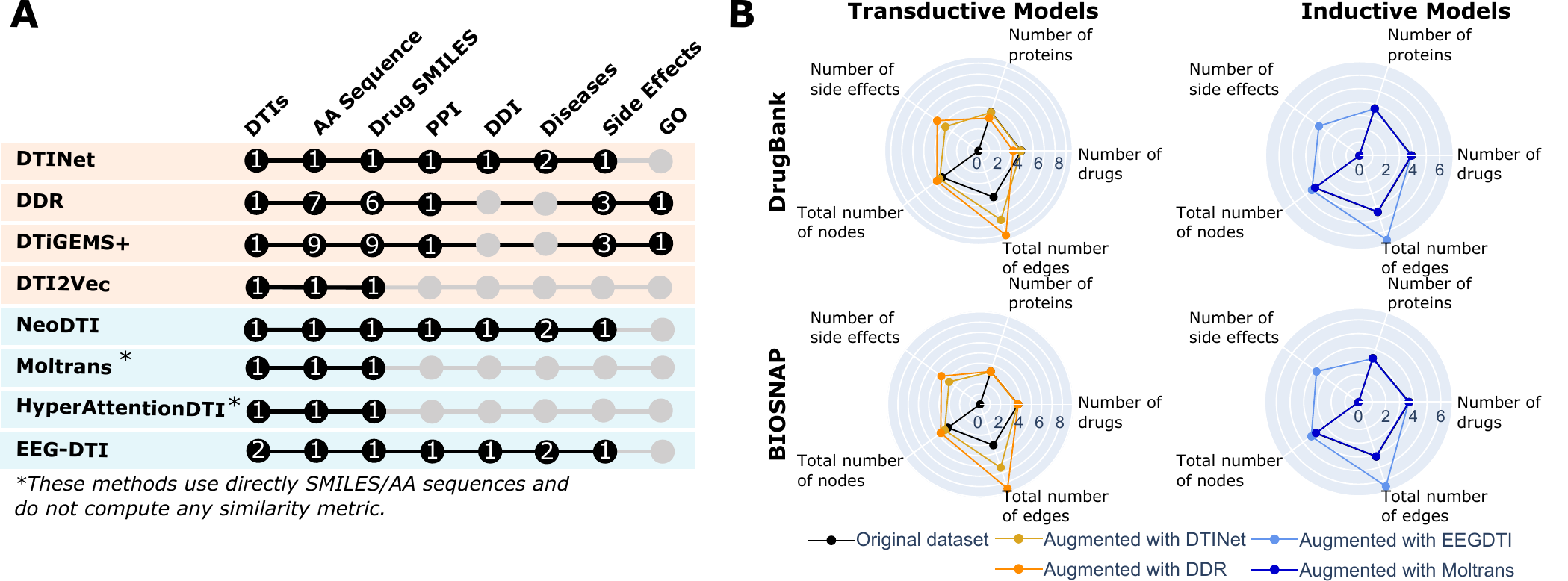}
\end{center}
\caption{\textbf{Analysis of required resources and network augmentation. }\textbf{A}. Number of different side information matrices used by each model. \textbf{B}. Radar plots depicting the original and modified number of nodes and edges (log10) for DrugBank and BIOSNAP, when used as input for different models. 
}
\label{fig:augmented-datasets}
\end{figure}

\textbf{Data requirements for current DTI prediction methods hinder their applicability.} DTI prediction methods typically augment the above-mentioned DTI datasets to include additional information beyond DTIs, such as protein-protein interactions or side-effect-drug associations. For example, methods like DTINet, DDR or DTIGEMS+ \cite{luo2017dtinet,olayan2018ddr,thafar2020dtigems} require collecting information from several complex data sources, such as Side Effects from SIDER \citep{kuhn2016sider}, or Diseases from CTD \citep{davis2021ctd} which hinders their usability (Fig.~\ref{fig:augmented-datasets}-A, Appx. Note \ref{sec:methods:datasets}). 
Generating these heterogeneous networks requires accessing information that may not be always readily available due to the inconsistency of identifiers across databases. Further, the absence of drug-target pairs in any required additional matrix precludes some models from including such pairs in the final graph. Indeed, the original number of proteins and drugs when using DrugBank, BindingDB, and NR datasets are considerably shrunk when used in high-demanding side-information models, losing up to 82\% of the drug nodes and 72\% of the protein nodes for DDR in DrugBank (Fig.~\ref{fig:augmented-datasets}-B). In approaches that require less demanding side information, such as Moltrans, the dataset size is maintained.

\textbf{A publicly available resource of augmented DTI datasets.} To enable robust benchmarking across DTI prediction models  with different augmented datasets, we built an augmented version of the most-used DTI datasets, including the gold standard. We computed all complementary matrices with the latest data releases required by every evaluated DTI prediction model (Fig.~\ref{fig:augmented-datasets}-A, Supp. Note \ref{sec:methods:datasets}). 

\begin{table}[htb]
\caption{\textbf{AUC (mean and std) for the evaluated DTI prediction models.} AUC shown for 
the eight state-of-the-art models (first four are transductive, second four are inductive) for every dataset. 
}
\label{tab:res-default}
\begin{center}
\resizebox{\columnwidth}{!}{%
\begin{tabular}{lcccccccc}
\hline\\
\textbf{Method} & \textbf{DrugBank} & \textbf{BIOSNAP}  & \textbf{BindingDB} & \textbf{DAVIS} & \begin{tabular}[c]{@{}c@{}}\textbf{Yamanishi}\\ \textbf{E}\end{tabular}  & \begin{tabular}[c]{@{}c@{}}\textbf{Yamanishi}\\ \textbf{IC}\end{tabular}  & \begin{tabular}[c]{@{}c@{}}\textbf{Yamanishi}\\ \textbf{GPCR}\end{tabular}  & \begin{tabular}[c]{@{}c@{}}\textbf{Yamanishi}\\ \textbf{NR}\end{tabular}  \\
\\ \hline \\
\rowcolor[HTML]{eaeaea} 
DTINet & 0.815 $\pm$ 0.000 & 0.856 $\pm$ 0.001 & 0.853 $\pm$ 0.005 & 0.813 $\pm$ 0.008 & 0.904 $\pm$ 0.004 & 0.737 $\pm$ 0.010 & 0.792 $\pm$ 0.003 & 0.803 $\pm$ 0.022 \\ 
DDR & OOT & OOT & OOT & 0.934 $\pm$ 0.020 & 0.975  $\pm$ 0.008 & 0.986 $\pm$ 0.006  & 0.957  $\pm$ 0.023 & 0.917 $\pm$ 0.049 \\
\rowcolor[HTML]{eaeaea} 
DTi-GEMS & OOR & OOR & 0.922 $\pm$ 0.034 & 0.843 $\pm$ 0.160 & 0.968 $\pm$ 0.230 & 0.973 $\pm$ 0.080 & 0.777 $\pm$ 0.107 & 0.935 $\pm$ 0.120 \\
DTi2Vec & OOR & OOR & 1.000 $\pm$ 0.004 & 0.862 $\pm$ 0.001 & 0.999 $\pm$ 0.009 & 0.996 $\pm$ 0.007 & 0.992 $\pm$ 0.004 & 0.968 $\pm$ 0.007 \vspace{0.15cm} \\ 
\hdashline \vspace{-0.05cm} \\ \rowcolor[HTML]{eaeaea}
NeoDTI & OOT & OOT & OOT & OOT & OOT & 0.974 $\pm$ 0.006 & 0.924 $\pm$ 0.021 & 0.825 $\pm$ 0.094 \\
Moltrans & 0.798 $\pm$ 0.008 & 0.792 $\pm$ 0.008 & 0.896 $\pm$ 0.009 & 0.693 $\pm$ 0.013 & 0.883 $\pm$ 0.011 & 0.875 $\pm$ 0.012 & 0.737 $\pm$ 0.037 & 0.544 $\pm$ 0.073 \\
\rowcolor[HTML]{eaeaea} 
HyperAttentionDTI & 0.862 $\pm$ 0.003 & 0.862 $\pm$ 0.003 & 0.954 $\pm$ 0.003 & 0.738 $\pm$ 0.003 & 0.953 $\pm$ 0.002 & 0.949 $\pm$ 0.006 & 0.804 $\pm$ 0.003 & 0.459 $\pm$ 0.001 \\ 
EEG-DTI & 0.889 $\pm$ 0.005 & 0.902 $\pm$ 0.010 & 0.893 $\pm$ 0.021 & 0.741 $\pm$ 0.034 & 0.951 $\pm$ 0.009 & 0.940 $\pm$ 0.017 & 0.870 $\pm$ 0.041 & 0.765 $\pm$ 0.084\\ \hline
\end{tabular}}
\begin{flushleft}
    \footnotesize{\tiny OOT and OOR mean \textit{out of time} ($\ge$ 7 days) and \textit{out of RAM} ($\ge$ 250 GBs), respectively.}
\end{flushleft}
\end{center}
\end{table}


\textbf{Evaluation of current DTI prediction models.} Using these augmented datasets, we then evaluated the above-mentioned methods following the originally proposed evaluation benchmark (Appx. Table \ref{tab:model-comparison}, default splitting column). Transductive models yielded significantly better AUC and AUPRC results (Table~\ref{tab:res-default}, Appx. Table \ref{suptab:auprc-default}, \ref{suptab:time-consumption}). However, except for DTINet, they did not converge on the two largest networks (DrugBank and BIOSNAP), potentially due to their large DTI network size, which gets further augmented with the needed additional matrices. On the other hand, inductive models  such as Moltrans and  HyperAttentionDTI obtained low AUCs in the smallest network, NR, indicating that the size of the network may be hampering the model learning capabilities.

\section{Graph ablation studies are crucial for fair and robust benchmarking
}\label{sec:splitting}

Since transductive methodologies can present data leakage during feature generation \cite{park2012flaws}, we hypothesized that the significant AUC discrepancies among inductive and transductive methods shown in the previous section could be a consequence of data leakage. This would indicate that the high AUC values achieved by these methods are not representative of their true ability to predict interactions as the model evaluation setting artificially raises the performance. As such, transductive methodologies should be carefully applied when designing DTI prediction approaches as information used in generating node embeddings 
is shared with the test set. Therefore, it becomes remarkably challenging to establish fair benchmarking guidelines as one must ensure that each model leverages the data in its intended way while ensuring fairness in cross-model comparisons.

\textbf{Previously proposed ablation studies.} To start addressing these challenges, recent reviews have proposed various scenarios regarding how drugs and proteins should be distributed among train-test splits \cite{pahikkala2014,olayan2018ddr}: $S_p$, where drugs and proteins are shared within train-test splits, $S_d$, where only proteins are shared, and $S_t$, where only drugs are shared. Despite these efforts to provide a homogeneous benchmarking practice, current DTI prediction models are typically evaluated considerably differently by their authors which makes it challenging to compare the performance across DTI prediction models (Appx. Table \ref{tab:model-comparison}). 

\begin{figure}[ht]
\begin{center}
\includegraphics[width=\linewidth]{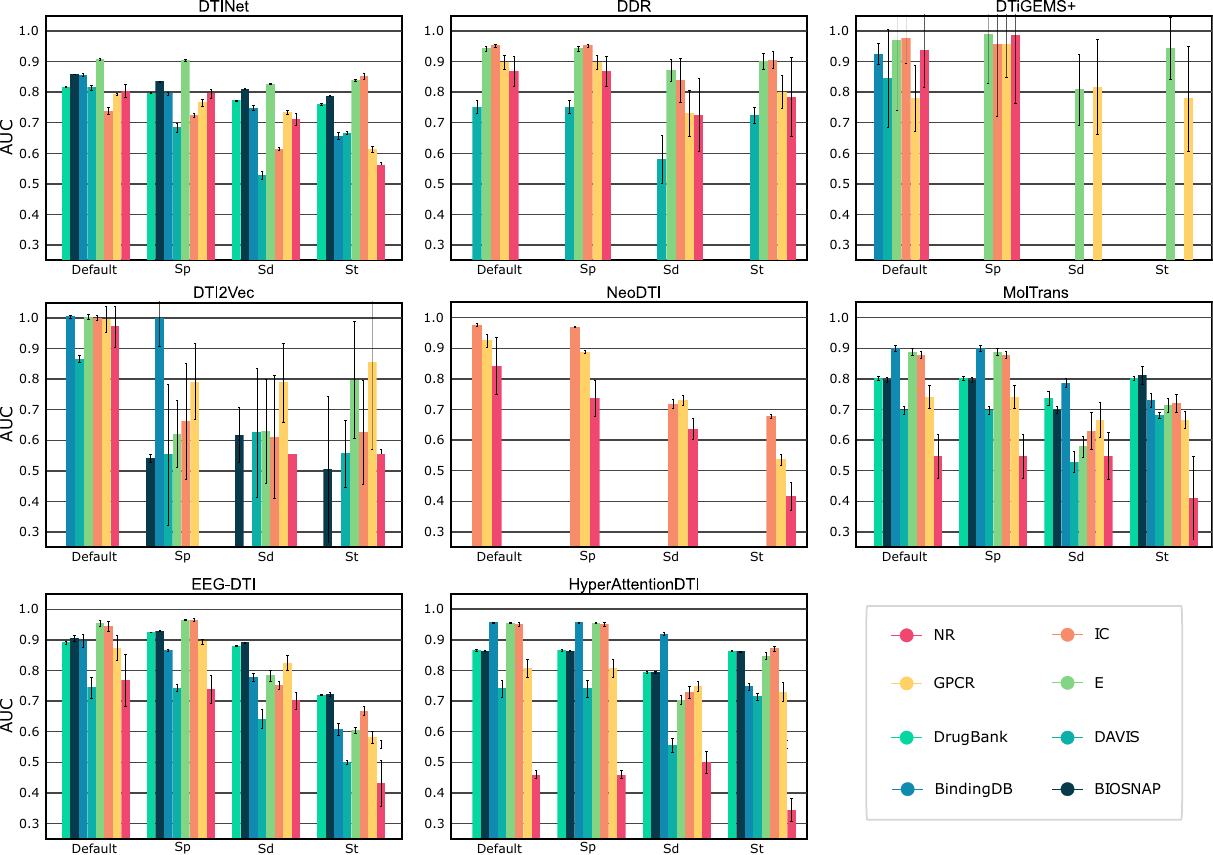}
\end{center}
\caption{\textbf{DTI prediction models benchmarking.} AUC results for each dataset and model (see Appx. Tables \ref{suptab:results_split_Sp}, \ref{suptab:results_split_Sd}, \ref{suptab:results_split_St} for AUPRC). Results correspond to an average of 5-Times 10-Fold Cross Validation. DDR, DTiGEMS+, MolTrans and HyperAttentionDTI models use $S_p$ as the default split.} 
\label{fig:splitting}
\end{figure}

\textbf{Ablation studies on state-of-the-art DTI prediction models.} We next performed an evaluation of the above-mentioned DTI prediction models using the different split scenarios for all the generated augmented DTI datasets (see previous section). As hypothesized, the overall results showed that the transductive models suffered a higher loss of AUC compared to their default split while the inductive models achieved a similar accuracy (Fig.~\ref{fig:splitting}). Interestingly, the transductive approach DTi2Vec, which exhibited exceptionally high performance during the default run by attaining a 0.999 AUC score on the Yamananishi's Enzyme dataset, yielded a performance decline to 0.61 when utilizing the \textit{$S_p$} split. The same behavior was observed when comparing the AUC scores within the models; \textit{$S_d$} and \textit{$S_t$} accuracies tended to be lower than those for \textit{$S_p$} (Fig.~\ref{fig:splitting}). 
Part of this variability appears to depend on the evaluated model, as demonstrated by the performance of different models on the DAVIS dataset. Specifically, the AUC scores were consistently higher for both \textit{$S_p$} and \textit{$S_t$} splits and lower for the \textit{$S_d$} split across most models. However, significant performance discrepancies were observed for the EEG-DTI and DTi2Vec models.

Furthermore, smaller datasets (e.g., Yamanishi’s NR) yielded a significant decrease of accuracy. This phenomenon was especially notable in DTi2Vec and DTiGEMS+, where removing edges hindered training on \textit{$S_p$} and aggravated both training and testing in \textit{$S_d$} and \textit{$S_t$}, as only very few edges remain in the network. A similar trend was observed in inductive methods like Moltrans and HyperAttentionDTI. While these methods can still be trained, their AUC score falls below 0.5, suggesting an inability to perform the task effectively, thereby emphasizing the need for larger networks. 

\textbf{Summary.} The performed ablation studies based on different train-test types of splits offer a more complete and realistic evaluation of DTI prediction approaches and suggest a lack of generalization capabilities of current transductive methodologies. 

\section{Train-to-Test data leakage in transductive DTI models prevents them from generalizing and yields inflated performance}\label{sec:node2vec}

We noticed that the best-performing transductive models, DTiGEMS+ and DTI2Vec,  shared the use of \textit{node2vec} (N2V) to generate the node embeddings for the DTI network \cite{grover2016node2vec}. As the node embeddings in N2V are built by local neighborhood visits within the network, it requires to be rerun whenever a new sample is included in the dataset. Thus, when used in DTI prediction methodologies, if the DTI network embedding using N2V occurs prior to the network splitting, it can promote data leakage issues when performing traditional train/test folds evaluation.

\begin{figure}[ht]
\begin{center}
\includegraphics[width=1\linewidth]{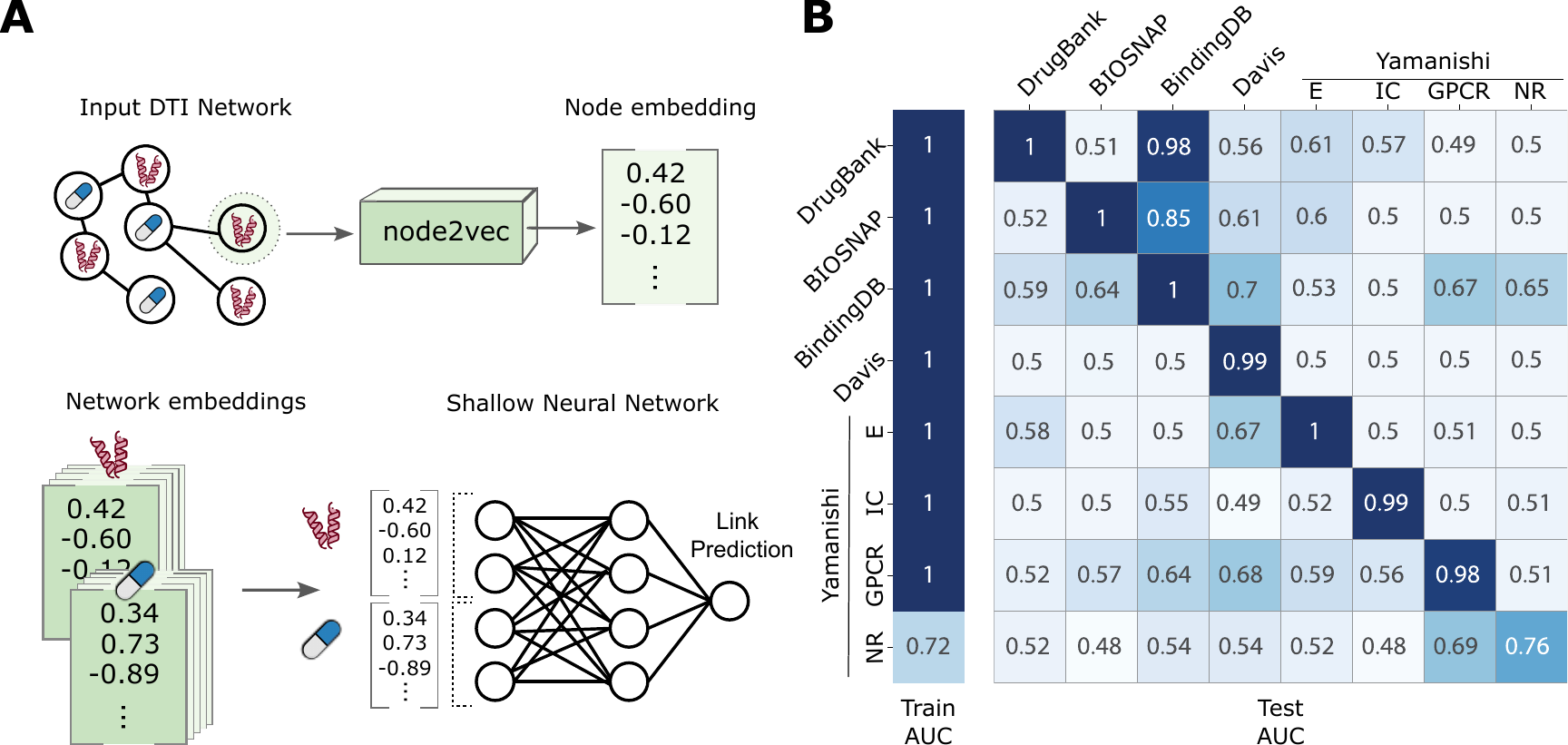}
\end{center}
\caption{\textbf{Evaluation of the designed \textit{node2vec}-based DTI prediction model.} \textbf{A}. Baseline classifier schematic. N2V embeddings are generated solely from the DTI network and drug-target pairs are fed into a SNN classifier. \textbf{B}. AUC matrix built by training on each dataset (left) and testing on another (right) (see Appx. Note~\ref{sec:methods_n2v}).}
\label{fig:node2vec}
\end{figure}

\textbf{Design of a baseline transductive model.} To delve into this potential data leakage, we designed a baseline model (Fig.~\ref{fig:node2vec}-A) based on N2V followed by a shallow neural network (SNN). We then performed a grid-search over multiple model parameters (Appx. Table \ref{supptable:n2v_gridsearch}) following a train/val/test evaluation setup,  across all assessed datasets. We report the test AUC scores, and found that for all datasets there is an optimal embedding dimension for which the variance of the AUC scores is minimized, while maintaining a high AUC score (Appx. Fig.~\ref{suppfig:n2v_embdim}). This variance is mostly influenced by the size of the datasets, as it decreases for larger networks. 

\textbf{Assessment of generalization capabilities of transductive DTI prediction models.} We trained and tested the baseline model on different DTI networks, generating an AUC matrix (Fig.~\ref{fig:node2vec}-B). The matrix's diagonal represents the test AUC when both the training and testing data come from the same dataset, aligning with the benchmarking process for the DTI prediction models (Table \ref{tab:res-default}). These results raised concerns about their reliability, as they consistently outperformed other evaluated methods across all datasets without leveraging any additional biological information.

Furthermore, this near-perfect performance drastically compares with the poor performance of the upper and lower triangles (where the train and test were constructed using different datasets, Appx. Note \ref{sec:methods_n2v}). This behavior aligns with what we observed for transductive models in the previous section and reaffirms that their inflated performance (Table~\ref{tab:res-default}) may be due to data leakage, as information from the test fold is present on the node's embeddings used in the train fold. Also, this analysis reveals one major drawback of the N2V approach for building DTI prediction models: the embedding process generally yields considerably different node embeddings for every network, hindering its capability to translate to unseen data. This also complicates generating embeddings on training and test folds separately to prevent data leakage, as the change in the graph topology produced by the splitting plus the transductive nature of N2V will heavily influence the generated embeddings.

\textbf{Summary.} The conclusions drawn from our baseline model can be transferable to 
transductive models such as the path-category-based technique in DDR, or the network diffusion algorithm (random walk with restart) in DTINet, which reduces the confidence of their obtained results. These findings also promote the adoption of inductive models for DTI prediction tasks, as the predictive models they construct during training enable testing on unseen graphs, mitigating the risk of data leakage and rendering them more suitable for a production environment.

\section{Protein structure-based metric leads to improved accuracy over currently used random subsampling
}\label{sec:subsampling}

When training DTI prediction models, the choice of positive and negative DTIs is still a challenging task. The sparse nature of current DTI networks, when used for classification tasks, yields very unbalanced datasets. The true edges are few and the negative edges, which are defined by all other possible connections, are orders of magnitude greater in number (see sparsity ratio in Appx. Table~\ref{tab:statistics}). 
%
Random subsampling is the preferred method to balance negative and positive edges. However, this can hamper the prediction task, as it is likely that the model is not trained on hard-to-classify negative samples. To address this issue, we propose a novel way to subsample negative DTIs that relies on the target's structural information to find hard-to-classify negative DTIs (Fig.~\ref{fig:rmsd_subsampling}-A).

\textbf{Identifying informative negative edges via Root Mean Square Deviation (RMSD) between backbone alpha carbons of two proteins.} Since evolution preserves protein structure more than the sequence itself \citep{siltberg2011evolution}, we consider those drug-target pairs with potential structural interaction to be plausible (uncovered) edges, measured using the RMSD between backbone C-alpha of two proteins. 
Hence, this metric ranks, for each positive DTI, all the negative pairs containing the same drug according to the structural similarity, which enabled us to identify hard-to-classify samples (negative pairs with low RMSD) and select high-quality negatives (negative pairs with RMSD within a defined window) (see Appx. Note~\ref{sec:rmsd-method} for further details on protein structure and RMSD calculations). 
Thus, the proposed subsampling scheme consists of two differentiated steps: the proposed ranking of edges via the RMSD metric,  and a selection of the negative edges for training. 

\textbf{Assessment of the proposed negative sampling criteria.} 
Since the analysis of subsampling methods becomes particularly relevant when dealing with larger DTI datasets, 
we tested the proposed sampling methodology on the largest datasets: BIOSNAP and BindingDB. For the evaluation methods, we discarded approaches based on N2V because of the aforementioned potential lack of generalization, as well as slow or hard-to-evaluate methodologies. From the remaining models, we chose MolTrans and HyperAttentionDTI, because of their inductive nature, their easiness of use, and their good performance in our previous analysis (Fig.~\ref{fig:rmsd_subsampling}-B, Appx. Fig.~\ref{suppfig:rmsd_hyperattentiondti}). 

The RMSD criteria helped both models to generalize and obtain more robust results, for almost every selected window (see Fig.~\ref{fig:rmsd_subsampling} and Appx. Note \ref{sec:rmsd-method})  across datasets and methodologies. Further, the AUC tends to decrease as we increase the RMSD, i.e., relax the similarity criteria. This is an expected behavior as including more easy-to-classify negative pairs into the folds can potentially decrease the total AUC score when tested on hard-to-classify DTI pairs. In summary, these results emphasize the importance of considering biologically driven criteria for future DTI prediction models' design and show the potential of the proposed negative-edge selection process for increasing the chances of uncovering novel DTIs.

\textbf{All previous analysis have been integrated into the python package \textit{GraphGuest}.} It enables 1) easy ablation studies of input DTI graphs, 2) computation of the RMSD-based score for negative edge selection, and 3) seamless access to the generated augmented DTI datasets (Appx. Note \ref{sec:code_availability}).

\section{RMSD-based subsampling enhances models capabilities for identifying novel interactions}
The utilization of RMSD-based selection for negative edges leads to an improved AUC, potentially facilitating the discovery of novel DTI interactions. 
To further investigate this hypothesis, we examined the excluded negative DTIs (2.5 to 5 \AA) within the largest selected network, BIOSNAP, using Moltrans 
trained on the highest yield AUC window (5 to 6 \AA). From the later, we specifically chose a set of DTIs that showed a high confident prediction when using RMSD, across five independent runs. From those, we selected EGFR and GSK3$\upbeta$ targets, as their inactivation elicited an antiproliferative effect, hence facilitating \textit{in vitro} validation. Finally, we compared the RMSD-based probabilities with the random subsampling ones, finding that the proposed metric consistently reported higher and more reliable positive predictions (Fig.~\ref{fig:rmsd_subsampling}-B).

\subsection{In vitro validation of newly uncovered drug-target interactions.}

We conducted \textit{in vitro} validation experiments to further assess their accuracy. The selected DTIs for subsequent validation were EGFR with 2-Amino-6-cyclohexylmethoxy-8-isopropyl-9H-purine (N69) and GSK3$\upbeta$ with carbinoxamine due to the readily availability of both reactives at our lab. 

\begin{figure}[ht]
\begin{center}
\includegraphics[width=0.93\linewidth]{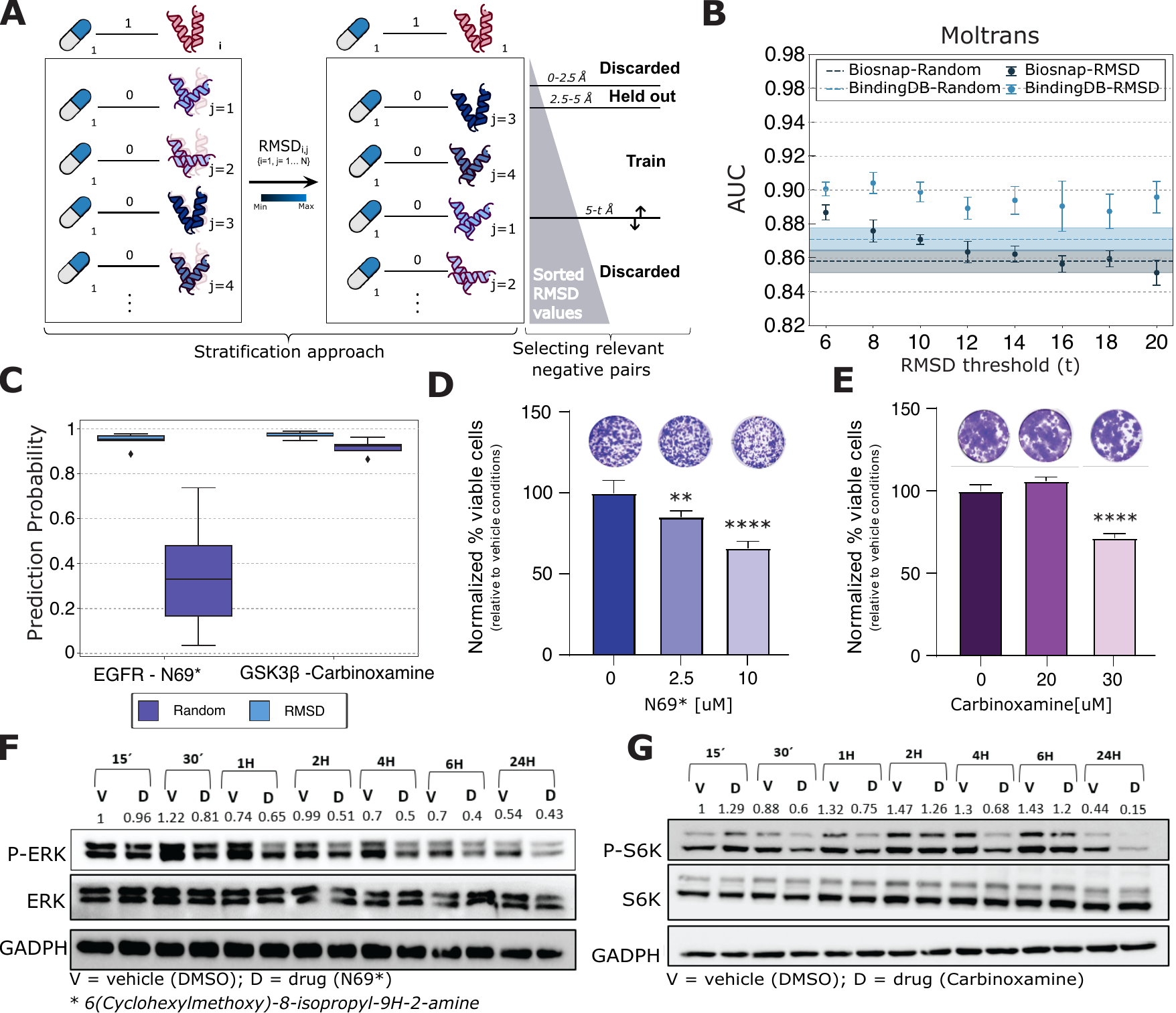}
\end{center}
\caption{\textbf{Biological criteria as an alternative to random subsampling}. \textbf{A}. Alternative
criteria proposed for negative subsampling based on RMSD protein structure comparison. \textbf{B}. AUCs of the test split for the RMSD-based (threshold from 6 to 20 \AA) and random subsampling tehcniques when using Moltrans model, on Biosnap and BindingDB datasets. \textbf{C}. Prediction probabilities (across five independent runs) of Moltrans for interactions EGFR - N69, and GSKB - Carbinoxamine, when using random-based and RMSD-based subsampling.
\textbf{D-E.} Percentage of cell viability and representative images of crystal violet-stained cells for (\textbf{D}) KRASG12D pancreatic cell line (HPAFII) after three days of treatment with N69  (0-10 $\mu M$) and (\textbf{E}) KRASG12C LUAD cell (H1792) after five days of treatmentwith Carbinoxamine (0-30 $\mu M$).
\textbf{F-G.} Protein expression at different time points after N69 10$\mu M$ (\textbf{F}) or Carbinoxamine 30µM (\textbf{G}) drugs, or DMSO vehicle treatment.
Relative P-Erk $\frac{1}{2}$ (\textbf{F})  and P-S6 Kinase (\textbf{G}) densitometry quantification is shown. 25$\mu g$ of protein were loaded per sample. 
\textit{GAPDH is shown as loading control.}
 }
\label{fig:rmsd_subsampling}
\end{figure}

\textbf{In vitro validation of EGFR-N69 interaction.} We used a cell line with active basal EGFR levels. We first assessed cell viability in the presence of the predicted drug. Cell viability decreased upon N69 treatment, particularly on the highest concentration of the compound, suggesting a potential interaction (Fig.~\ref{fig:rmsd_subsampling}-C). 
To understand if this effect was specific to the inactivation of the EGFR pathway, we examined the activation level of downstream effectors within the target pathway (Appx. Fig.~\ref{suppfig:EGF-ERK-pathway}). 
Specifically, we examined the activation state of the RAF-MEK-ERK pathway in HPAFII cells using antibodies specific to phospho-ERK1/2. We found that while the protein levels of ERK, as well as those of a protein loading control (GAPDH) remained unchanged in the presence of the drug, the phosphorylated active version (P-ERK) decreased as early as 30 min post-treatment. Thus, the reduction in P-ERK may be a consequence of the observed decrease in cell viability (Fig.~\ref{fig:rmsd_subsampling}-E) \cite{EGFR_validation}.

\textbf{In vitro validation of GSK3$\upbeta$-Carbinoxamine interaction.} Similarly, we tested the GSK3$\upbeta$-Carbinoxamine interaction, revealing a decrease in cell viability at a 30 $\mu M$ concentration (Fig.~\ref{fig:rmsd_subsampling}-D). When looking into the downstream proteins reported to mediate GSK3$\upbeta$ signaling (Appx. Fig.~\ref{suppfig:insulin-pathway}), we found that the levels of phosphorylated S6 kinase, a protein involved in protein synthesis when activated through phosphorylation, were reduced 30 minutes after Carbinoxamine treatment while its non-phosphorylated version and the loading protein control remained unchanged (Fig.~\ref{fig:rmsd_subsampling}-F) \cite{zhang2006s6k1}. 

\textbf{Summary.} These observations suggest that the antiproliferative effect caused by N69 and Carbinoxamine may indeed be related to inactivation of protein effectors downstream of the predicted targets EGFR and GSK3$\upbeta$. Thus, the RMSD selection method may increase the likelihood of discovering positive DTIs compared to random subsampling. 

\section*{Discussion}\label{sec12}


Previous in-silico drug repurposing methodologies have often required high-demanding additional information, exhibited significant disparities in their evaluation framework, and employed structurally distinct learning architectures, which has resulted in a lack of a standardized benchmarking approach to determine the most suitable model.
In this study, we started by assessing currently used datasets in DTI prediction problems, and generated a valuable resource of augmented DTI datasets that will enable accessible and robust future benchmarking of DTI prediction models. Using this newly generated data resource, we benchmarked diverse drug repurposing models by first using the traditional approach and then following graph-aware train-test splitting techniques. The latter revealed that methods employing transductive feature generation exhibited over-performance. This motivated further assessment of transductive approaches, which allowed uncovering data leakage issues that could be avoided if using inductive approaches.

To improve the predictive capabilities of inductive DTI models we proposed a subsampling method based on structural differences across proteins. This revealed improved accuracy when compared  with traditional random subsampling, increasing the reliability of uncovering novel DTIs. Importantly, we then performed \textit{in vitro} validation suggesting a direct interaction between drugs and protein targets leading to potential pathway inactivation, as revealed by variations in the activation levels of canonical downstream effectors of the targeted proteins. 

\textbf{Conclusion.} This study emphasizes the significance of larger and diverse DTI databases, accessible drug repurposing models, data-leakage-free evaluation, and biologically driven subsampling techniques. It also presents the \textit{GraphGuest} python package that will ease the design of drug repurposing approaches.

\setcounter{section}{0}
\section*{Appendix}
\section*{Appendix Notes}
\setcounter{subsection}{0} 
\renewcommand\thesubsection{\arabic{subsection}}

\subsection{Datasets}\label{sec:methods:datasets}
Along the work, multiple DTI networks have been evaluated. Here we briefly describe them all:
\begin{itemize}
    \item \textbf{DrugBank} \citep{wishart2006drugbank}. DTIs collected from DrugBank Database Release 5.1.9. It has undergone significant upgrades since its first release in 2006. 
    \item \textbf{BIOSNAP} \citep{zitnik2018biosnap}. Dataset created by Stanford Biomedical Network Dataset Collection. It contains proteins targeted by drugs on the U.S. market from DrugBank release 5.0.0 using MINER \citep{miner}.
    \item \textbf{BindingDB} \citep{liu2007bindingdb}. Database that consists of measured binding affinities, focusing on protein interactions with small molecules. The binarization of the dataset was done by considering interactions as positive if their Kd was lower than 30 units. Data was downloaded from Therapeutics Data Commons (TDC) \citep{tdcdatabase}.
    \item \textbf{DAVIS} \citep{davis2011comprehensive}. Dataset of kinase inhibitors to  kinases covering more than 80\% of the human catalytic protein kinome. The binarization of the dataset was done by considering as positive those interactions with a Kd lower than 30 units. Data downloaded from Therapeutics Data Commons (TDC) \citep{tdcdatabase}.
    \item \textbf{Yamanishi et al.} \citep{yamanishi2008prediction}.  It is composed of four subsets of different protein families: enzymes (E), ion channels (IC), G-protein-coupled receptors (GPCR) and nuclear receptors (NR). The Yamanishi dataset has been considered the gold standard dataset for DTI prediction and has been used in several published models \citep{zong2019,peng2021eegdti, zhang2022}. DTIs in this dataset come from  KEGG BRITE \citep{kanehisa2006kegg}, BRENDA \citep{schomburg2004brenda}, SuperTarget \citep{gunther2008supertarget} and DrugBank. Compounds with molecular weights lower than 100 are excluded from the dataset. In the Enzyme group all the ligands are inhibitors or activators, and co-factors are not included.
\end{itemize}

Also, complementary datasets were used for building the augmented networks:

\begin{itemize}
    \item \textbf{CTD} \cite{davis2021ctd}. Comparative Toxicogenomics Database for disease-drug and disease-protein associations.

    \item \textbf{DrugBank} \cite{wishart2006drugbank}. DrugBank Database can be used for extracting other information such as drug-drug interaction.

    \item \textbf{FDA} Adverse Event Reporting System (FAERS) \cite{faers}. The FAERS is a database that contains adverse event reports, medication error reports and product quality complaints resulting in adverse events that were submitted to FDA.

    \item \textbf{HPRD} Human Protein Reference Database \cite{keshava2009hprd} for human protein-protein interactions.

    \item \textbf{SIDER} Side Effect Resource Database \cite{kuhn2016sider} aggregates information from side effects 

\end{itemize}

Further, other databases have been used to change between identifier types, e.g., KEGG Drug ID to PubChem ID, such as STITCH \cite{kuhn2007stitch}, bioMART \cite{smedley2009biomart}, ChemBL \cite{gaulton2012chembl}.

\subsection{Related work}\label{sec:methods:models}

In what follows we briefly describe the selected state-of-the-art DTI models, where the first four are transductive and the second four are inductive. Note that referring to a model as transductive or inductive concerns the feature (embedding) generation process and not the prediction task itself.
\begin{itemize}
    \item \textbf{DTINet} \citep{luo2017dtinet}. DTINet considers a heterogeneous graph with four node types (drugs, proteins, side effects and diseases) and six edge types (DTIs, protein-protein interaction, drug-drug interaction, drug-disease association, protein-disease association, drug-side-effect association, plus similarity edges between drugs and proteins). After compact feature learning (based on a random walk with restart) on each drug and protein network, it calculates the best projection from one space onto another using a matrix completion method, and then infers interactions according to the proximity criterion. The matrices generated are known as “Luo Dataset”. 
    
    \item \textbf{DDR} \citep{olayan2018ddr}. DDR uses a heterogeneous graph built from known DTIs, multiple drug-drug similarities, and several protein-protein similarities. Firstly, DDR performs a pre-processing step where a subset of similarities is selected in a heuristic process to obtain an optimized combination of similarities. Then, DDR applies a non-linear similarity fusion method to combine different similarities. Finally, from these combined similarities, a path-category-based feature extraction method is applied, and these features are fed into a random forest model.
    
    \item \textbf{DTiGEMS+} \citep{thafar2020dtigems}. The information of the interaction within drugs and proteins coming from diverse matrices is selected and integrated to create a heterogeneous graph alongside the DTI information. Simultaneously, a second graph is created by applying \textit{node2vec} to the DTI graph, obtaining the features for each node and augmenting the interactions based on the similarity of the calculated features. Multiple paths are extracted from both graphs and feed to a supervised ML classifier after a feature selection process.
    
    \item \textbf{DTI2Vec} \citep{thafar2021dti2vec}. DTI2Vec stems from the previous and more complex model  DTiGEMS+, trying to improve the precision of the predictions while reducing the amount of side information needed. This method only uses the similarity matrices within drugs and proteins to increase the number of connections on the DTI network. The nodes of this augmented network are used as input to \textit{node2vec}, and the resulting embeddings are combined to create a feature vector and feed a classifier. 

    \item \textbf{NeoDTI} \citep{wan2018neodti}. NeoDTI aims to automatically learn a network topology-preserving node-level embedding to facilitate DTI prediction. First, neighborhood information aggregation and node embedding update processes ensure that each node within the heterogeneous network generates a new feature representation by integrating its neighborhood information with its own features. Then, they enforce the node embeddings to preserve the network topology, aiming to reconstruct the original individual networks. Finally, from these embeddings they extract the node features and use them for the DTI prediction.
    
    \item \textbf{MolTrans} \citep{huang2020moltrans}. MolTrans uses unlabeled data to decompose drugs and proteins into high-quality substructures. Then it creates an augmented embedding for each using a transformer and a map of interactions, allowing it to predict which substructures contribute most to the overall interaction. 

    \item \textbf{HyperAttentionDTI} \citep{zhao2021hyperatt}. HyperAttentionDTI embeds each character of the different sequences into vectors. Then the model makes use of an attention mechanism and convolutional neural networks (CNNs) to make DTI predictions. It models the complex non-covalent intermolecular interactions between atoms and amino acids using the attention mechanism.    
    
    \item \textbf{EEG-DTI} \citep{peng2021eegdti}. EEG-DTI considers a heterogeneous graph using the same type of dataset as DTINet. It first generates low-dimensional embeddings for drugs and proteins with three graph convolutional networks (GCN) layers and concatenates them separately. Then, it calculates their inner product to get a protein-drug score. 
\end{itemize}

Our \textbf{baseline classifier} (denoted as \textbf{N2V+NN}) is based on \textit{node2vec} to embed the DTI network so that it solely relies on the topology of the network. From the generated embeddings, positive edges and a random subsampling of negative edges are used to train and validate a 2-layer neural network $\Psi$. Being ${X} \in \mathbb{R}^{K \times 2d}$ the batched input matrix, and $\textbf{W}_{1} \in \mathbb{R}^{2d \times n}$ and $\textbf{W}_{2} \in \mathbb{R}^{n \times 1}$ the associated weight matrices, our model $\Psi$ will generate the output $h \in \mathbb{R}^{K \times 1}$ as:
\begin{equation*}
h = \sigma(\textbf{W}_{2} \cdot f(\textbf{W}_{1} \cdot X)), 
\end{equation*} 
where $d$ is the selected \textit{node2vec} embedding dimension for each node, $K$ is the number of samples per batch, $f$ is a ReLU activation function, $\sigma$ is a sigmoid activation function and $n$ is the number of neurons of the first layer. In order to solve the DTI classification problem, we use a loss that combines the sigmoid of the output layer and the binary cross entropy loss in a single function. This combination takes advantage of the log-sum-exp trick for numerical stability \citep{pytorch}. For each sample $x_k$ in a given batch ($k \in [1,K]$), the loss is given by: 
\begin{equation*} 
\quad l_k=-w_k\left[y_k \cdot \log h_k+\left(1-y_k\right) \cdot \log \left(1-h_k\right)\right], 
\end{equation*}

where $w_{k}$ is a manual rescaling weight, $y_{k}$ is the associated label for sample $x_k$, and $h_k$ is the model output for sample $x_k$. The final loss $L$ is then computed as the average of $(l_1, \ldots, l_K)$. We performed a train/validation/test (0.75, 0.15, 0.1) splitting prior performing a hyperparameter tuning, varying several architectures, loss functions, epochs and batch sizes to select the model with the highest test AUROC for every evaluated dataset (Appendix Table \ref{supptable:n2v_gridsearch}).

\subsection{Evaluation setup}\label{sec:set-up}
\subsubsection*{A fair evaluation scheme: graph embedding splitting approach}
The following evaluation scheme consisting of constructing three different train-test splits ($S_p$, $S_d$, and $S_t$) was used: 

\begin{itemize}
    \item \textbf{$S_p$} Related to pairs. Any protein or drug may appear both in the train and test set, but interactions cannot be duplicated in the two sets. 
    \item \textbf{$S_d$} Related to drug nodes. Drug nodes are not duplicated in the train and test set, i.e., a node evaluated during training does not appear in the test set. 
    \item \textbf{$S_t$} Related to targets. Protein nodes are not duplicated in the train and test set, each protein seen during training does not appear in the test set. 
\end{itemize}

If the model to be compared uses three splits (train/val/test), the criterion is applied the same way as if they were just two splits (train/test), but applying an extra split to the train fold, yielding train/val/test folds. Hence, train and validation will be evaluated together when verifying $S_p$, \textit{$S_d$} and \textit{$S_t$} splits. 

Note that most assessed models have not been evaluated on these splitting criteria, but perform a \textit{traditional} split. This consists of a random splitting of the DTI network without constraining the DTI distribution, which may lead to repetition of drug or protein nodes across folds. As the $S_p$, $S_d$ and $S_t$ splits impose certain constraints not assumed by the authors and may result in lower performance than what was initially reported, we also provide, for each model, the results following the originally proposed evaluation benchmark.

Furthermore, the $S_c$ split, related to a couple of different DTI networks that do not have common drugs nor proteins \citep{zhao2021hyperatt,pahikkala2014}, involves training a model initially on one dataset and then testing the trained model on another dataset.  This split can assist in assessing the methods' generalization capabilities, potentially revealing data leakage concerns. However, the limited reproducibility of most methods have complicated the application of this evaluation scheme to the evaluated ones. Nonetheless, we validated our hypothesis regarding \textit{node2vec}-based methods by applying this split to our baseline DTI classifier (see Appendix Note \ref{sec:methods:models}).

\subsubsection*{Building train and test splits for N2V evaluation}
\label{sec:methods_n2v}

In assessing the generalization capability of \textit{node2vec}-based drug repurposing models across multiple DTI networks, we evaluated the designed baseline model using train/test splits. First, node embeddings for each network were constructed individually using \textit{node2vec}. Next, for each network, a balanced dataset was created by selecting all positive pairs and randomly pairing them with negatives in a 1:1 ratio. Finally, the baseline model was trained on embeddings from one dataset and tested on a different one, yielding both train and test AUROC and AUPRC values.  When the same network is used for both training and testing (as shown in Fig. \ref{fig:node2vec}-B matrix's diagonal), the dataset was constructed as previously described, with a 70/30 train-test split.

\subsubsection*{Considering a biological driven criteria for negative subsampling}
\label{sec:rmsd-method}

Here now we describe the process of DTI stratification and hard-to-classify pairs selection. First, for each known DTI interaction (labeled as 1), we compute the RMSD between the selected protein and every other protein available in the data set. Then, in order to generate a balanced dataset, for each positive DTI, we select a protein to form a negative interaction, based on the computed RMSD between the known target and every other protein in the network. The selection is made by sorting the proteins' RMSD, and we will select or discard them based on three different windows. The first window ranges from 0 to 2.5 {\AA}, where proteins in this interval are discarded, as in this range we may include small structures or very simple proteins that align non-specifically to others. Proteins lying on the second window, from 2.5 to 5 {\AA}, are held out for validation, as they are very similar to the actual target but are labelled as 0, so they can generate false positives, potentially hinder the model's training. The third window, ranged from 5 to $t$ {\AA} ($t \in $ [6, 20]), will be the selected one as the train split, as is populated with proteins that are closely enough to the target to be a difficult train event, but different enough to assure the potentially true negativity of the data.

\subsubsection*{Tanimoto Similarity}
The pairwise drug similarity, calculated with Tanimoto metric was calculated in RDKit \citep{rdkit} creating fingerprints in default configuration using \textit{RDKFingerprint} (with 2048 bits) function.

\subsubsection*{Protein Structures and RMSD Calculation}
Protein structures were obtained from the PDB Database \citep{rcsbpdb} and Alpha Fold \citep{varadi2021afold,jumper2021afold}, considering X-Ray structures with resolution lower than 2 Å and a per-residue confidence score higher than 70 on average, respectively. 
The RMSD was calculated using an adapted script from PyMOL \citep{PyMOL}, considering superimposition mode and all objects aligned using the alpha carbons (C-alpha) of the backbone of the two proteins and the default configuration of 5 cycles. See Appendix Figs.~\ref{suppfig:s2_rmsd_suppl}, \ref{suppfig:s2_tani_suppl}, \ref{suppfig:s3_drugs_yamanishi_heatmap}, \ref{suppfig:s3_drugs_dbbd_heatmap},  for distribution of pairwise RMSD in all datasets.


\subsubsection*{Hardware}
All simulations were performed on a workstation with 64 cores Intel xeon gold 6130 2.1Ghz and 754Gb of RAM. A Quadro RTX 4000 GPU was also used, with driver version 460.67 and cuda 11.2. version.

\subsection{\textit{In vitro} validation}
\subsubsection*{Cell lines} 
Human mut KRAS (H1792) LUAD cell line and human mut KRAS (HPAF II) PDAC cell line were used. All these cell lines were obtained from American Type Culture Collection (ATCC)  and authenticated by the Genomics Unit at CIMA using Short Tandem Repeat profiling (AmpFLSTR Identifiler Plus PCR Amplification Kit). Human cells were grown according to ATCC specifications.

\subsubsection*{Reagents} 
Carbinoxamine maleato (PHR2802) was purchased from Merck and  6-(Cyclohexylmethoxy)-8-isopropyl-9H-purin-2-amine was synthesized and obtained from  Wuxi.

\subsubsection*{Western blotting}
Western blot methodology was performed as previously published \citep{vallejo2017}. For these experiments cells were treated with DMSO (vehicle, control conditions) or drug. In this last case, we used a final concentration of 30uM for Carbinoxamine and 10uM for 6-(Cyclohexylmethoxy)-8-isopropyl-9H-purin-2-amine. Antibodies used: GAPDH (1:5,000, ab9484, Abcam), ERK1/2 (1:1,000, \#9102, Cell Signaling Technology), p-ERK1/2 (1:1,000, \#9101, Cell Signaling Technology), p70S6K (1:1,000, \#2708, Cell Signaling Technology), p-p70S6K (1:1,000, \#9205, Cell Signaling Technology), EGFR (1:1,000, \#2232, Cell Signaling Technology), p-EGFR (1:1,000, \#2236, Cell Signalling Technology), GSK3$\upbeta$ (1:1,000, ab31826, Abcam), p-GSK3$\upbeta$ (1:1,000, \#9336, Cell Signalling Technology) and p-4E-BP1 (1:1,000, \#9451, Cell Signaling Technology).

\subsubsection*{Drug studies in vitro}
To determine the number of viable cells in proliferation and the potential cytotoxicity of drugs in cell lines, cells were seeded in triplicate into 96-well plates (range: 500 to 1,800 cells per well depending on the cell line). The next day, cells were cultured in the absence or presence of rising concentrations of single drugs (Carbinoxamine 0-30uM; 6-(Cyclohexylmethoxy)-8-isopropyl-9H-purin-2-amine 0-10uM) during 3 or 5 days. At these time points, remaining cells were fixed with 4\% formaldehyde (Panreac) for 15 minutes at RT, stained with crystal violet solution (Sigma-Aldrich) (1\% crystal violet in H2O) for 15 minutes and photographed using a digital scanner (EPSON Perfection v850 Pro). Relative growth was quantified by measuring absorbance at 570 nm in a spectrophotometer (SPECTROstar Nano – BMG Labtech) after extracting crystal violet from the stained cells using 20\% of acetic acid (Sigma).

\subsubsection*{Protein and Drug Annotation}
Proteins were annotated using Molecular Function Keywords from Uniprot \citep{uniprot2023} and drugs with Classyfire \citep{classyfire}. Annotated heatmaps were generated to check whether proteins cluster per molecular function and drugs by chemical classification.


\section*{Code availability}
\label{sec:code_availability}

Dockers for all evaluated models are available for in DockerHub. The repository containing all the developed tools and code, along with the \textit{GraphGuest} Python Package are available at \url{https://github.com/ubioinformat/GraphEmb}.

\section*{Acknowledgments}
We would like to thanks to Oier Azurmendi for his support along the development of the work.

\section*{Funding}
This work was supported by the following grants: DoD of the US - CDMR Programs [W81XWH-20-1-0262], Ramon y Cajal contracts [MCIN/AEI RYC2021-033127-I] [RYC2019-028578-I], DeepCTC [MCIN/AEI TED2021-131300B-I00], Gipuzkoa Fellows [2022-FELL-000003-01], the Spanish MCIN (PID2021-126718OA-I00), Fulbright Predoctoral Research Program [PS00342367], and  FEDER/MCIN - AEI (PID2020‐116344‐RB‐100/MCIN/AEI/10.13039/501100011033).

\section*{Competing interests}
The authors declared no competing interests.

\section*{Author contributions}
Conceptualization: J.F., G.S., U.V., I.O., O.G., and M.H.; methodology: J.F., G.S., U.V., I.O., and M.H.; software: J.F., G.S., U.V, and M.C.; formal analysis: J.F., G.S., U.V, and M.C.; investigation: J.F., G.S., U.V; validation: J.F., G.S., U.V., M.C and S.V.; supervision: I.O., S.V., O.G., and M.H.; writing-original draft: J.F., G.S., U.V., I.O., S.V. O.G., and M.H.; visualization: J.F., G.S., U.V.; writing-review \& editing: all authors.
\clearpage
\section*{Appendix Figures}

\counterwithin{figure}{section}
\setcounter{figure}{0} 

\renewcommand{\figurename}{Appendix Figure}
\renewcommand{\thefigure}{A\arabic{figure}}


\begin{figure}[H]
\begin{center}
\includegraphics[width=1\linewidth]{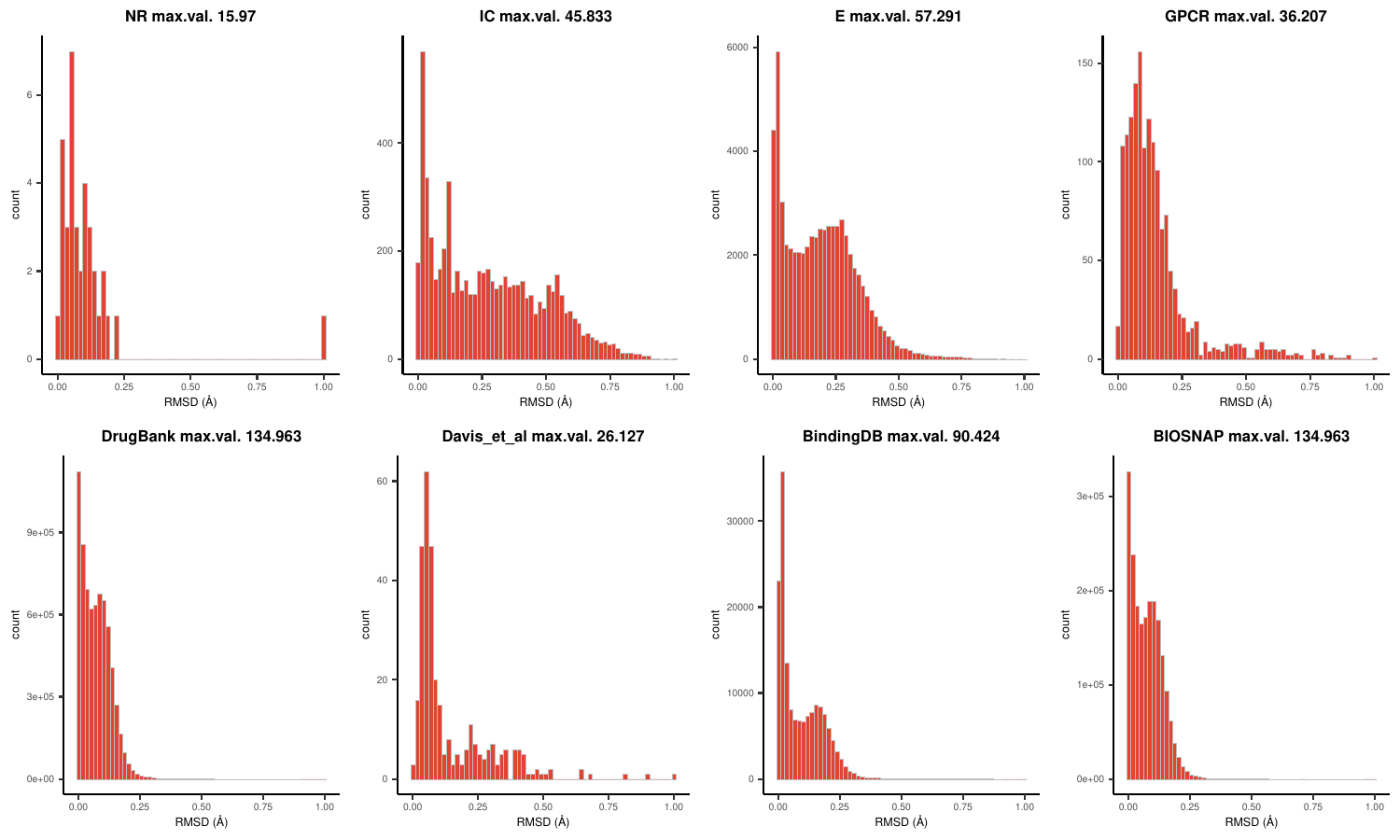}
\end{center}
\caption{Histograms of the distribution of the pairwise Root Mean Square Deviation (RMSD) of all proteins for each dataset, regardless of whether they come from PDB or AlphaFold.}
\label{suppfig:s2_rmsd_suppl}
\end{figure}

\begin{figure}[H]
\begin{center}
\includegraphics[width=1\linewidth]{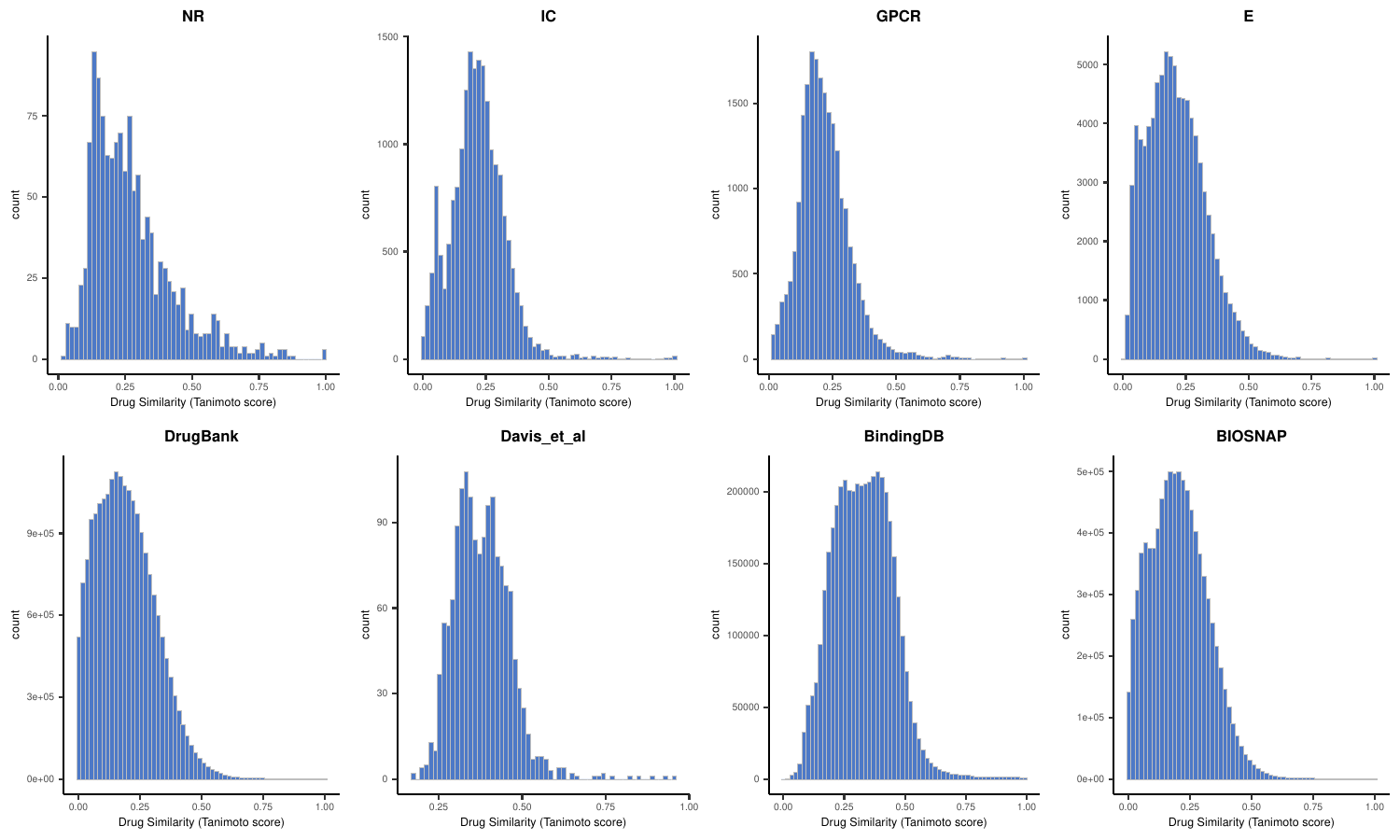}
\end{center}
\caption{Histograms of the distribution of Tanimoto score calculated pairwise over all drugs for each datataset. The distribution of chemical similitude shift to the left indicates that drugs are chemically diverse}
\label{suppfig:s2_tani_suppl}
\end{figure}


\begin{figure}[H]
\begin{center}
\includegraphics[width=1\linewidth]{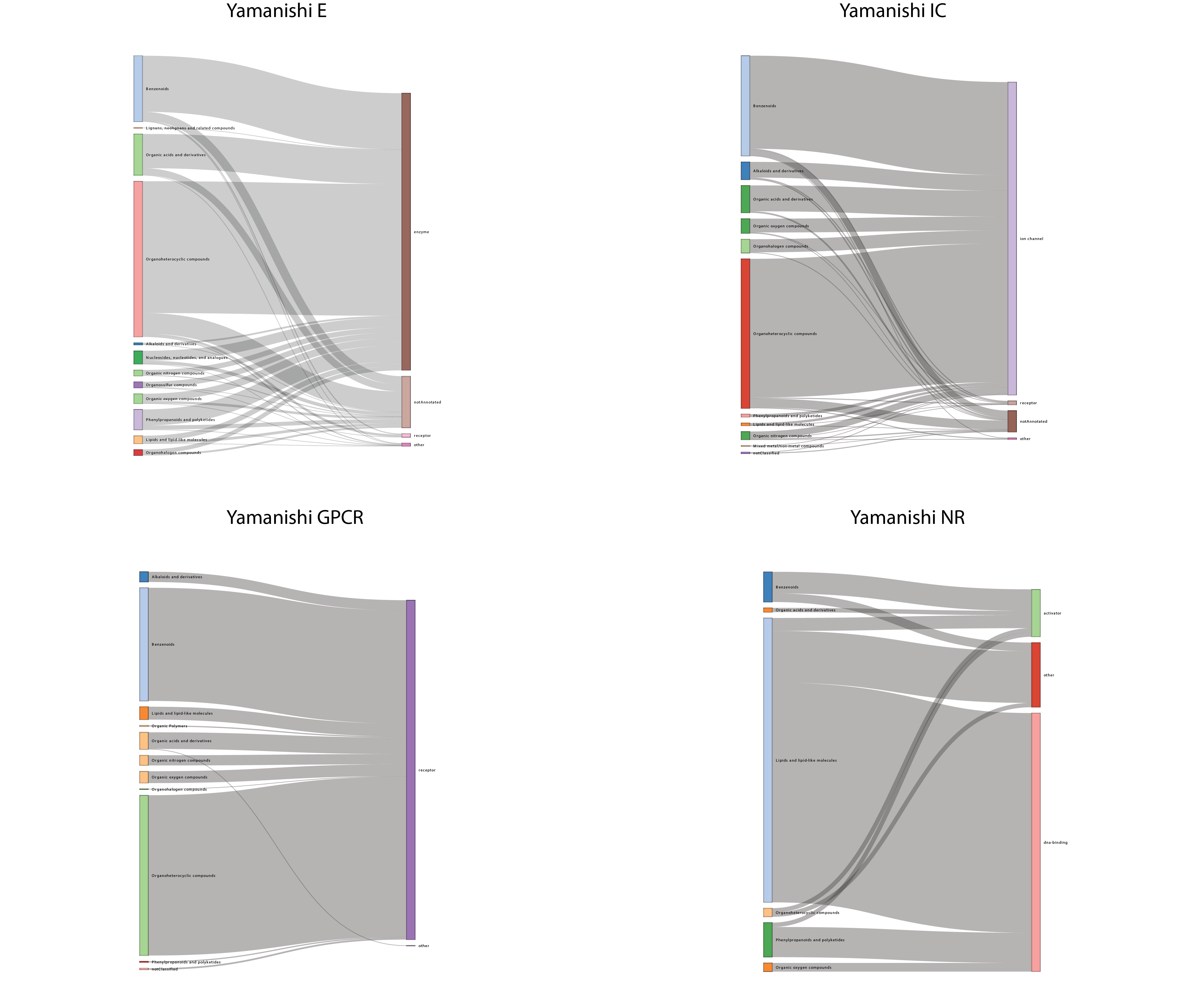}
\end{center}
\caption{Sankey plots of Yamanishi datasets. These Sankey plots connect the drug chemical category with the protein family, remarking that each connection is an existing DTI in the dataset.}
\label{suppfig:s5_sankey_yamanishi}
\end{figure}

\begin{figure}[H]
\begin{center}
\includegraphics[width=1\linewidth]{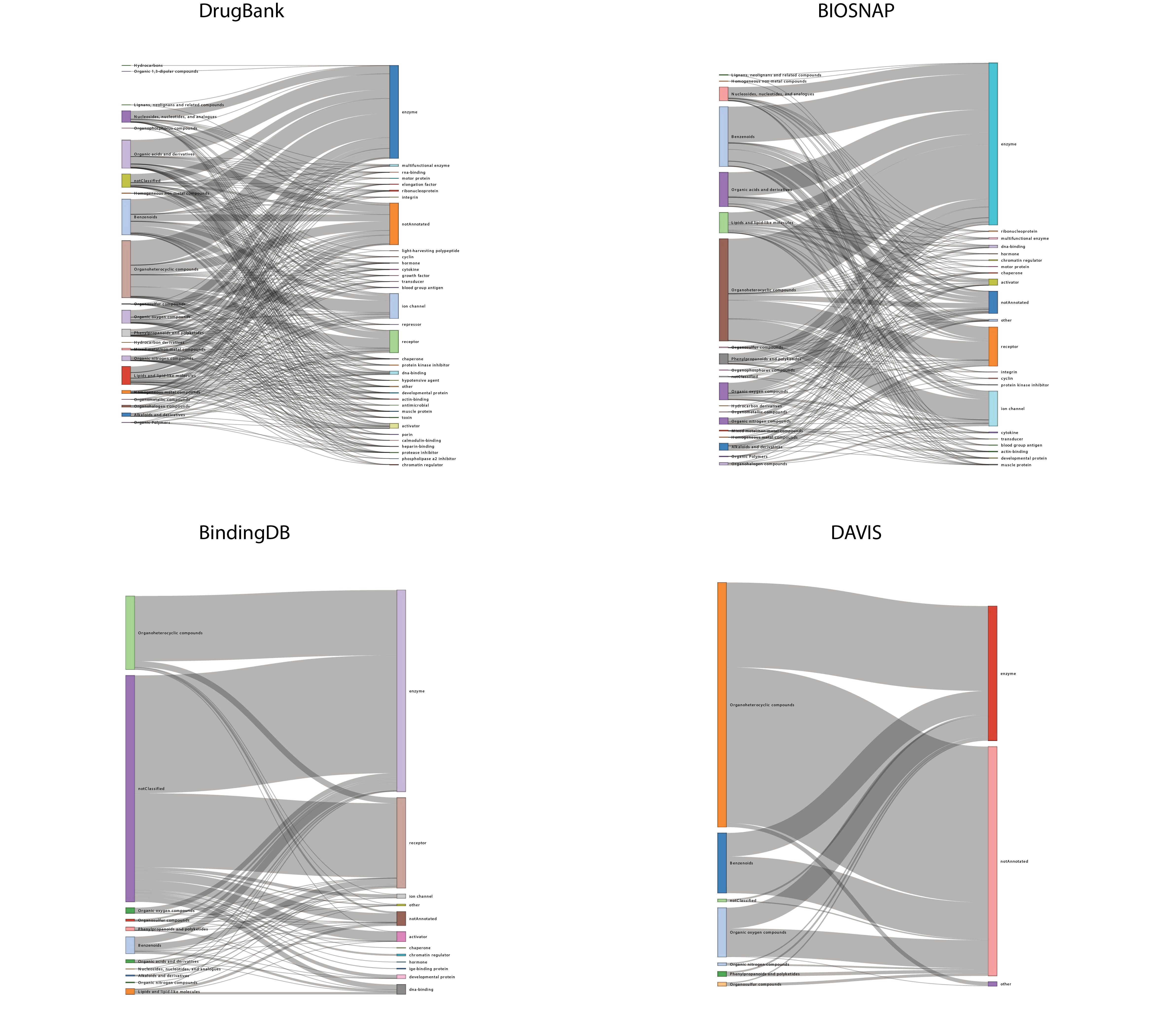}
\end{center}
\caption{Sankey plots of DrugBank, BIOSNAP, BindingDB and DAVIS datasets. These Sankey plots connect the drug chemical category with the protein family, remarking that each connection is an existing DTI in the dataset.}
\label{suppfig:s5_sankey_dbbd}
\end{figure}


\begin{figure}[H]
\begin{center}
\includegraphics[width=1\linewidth]{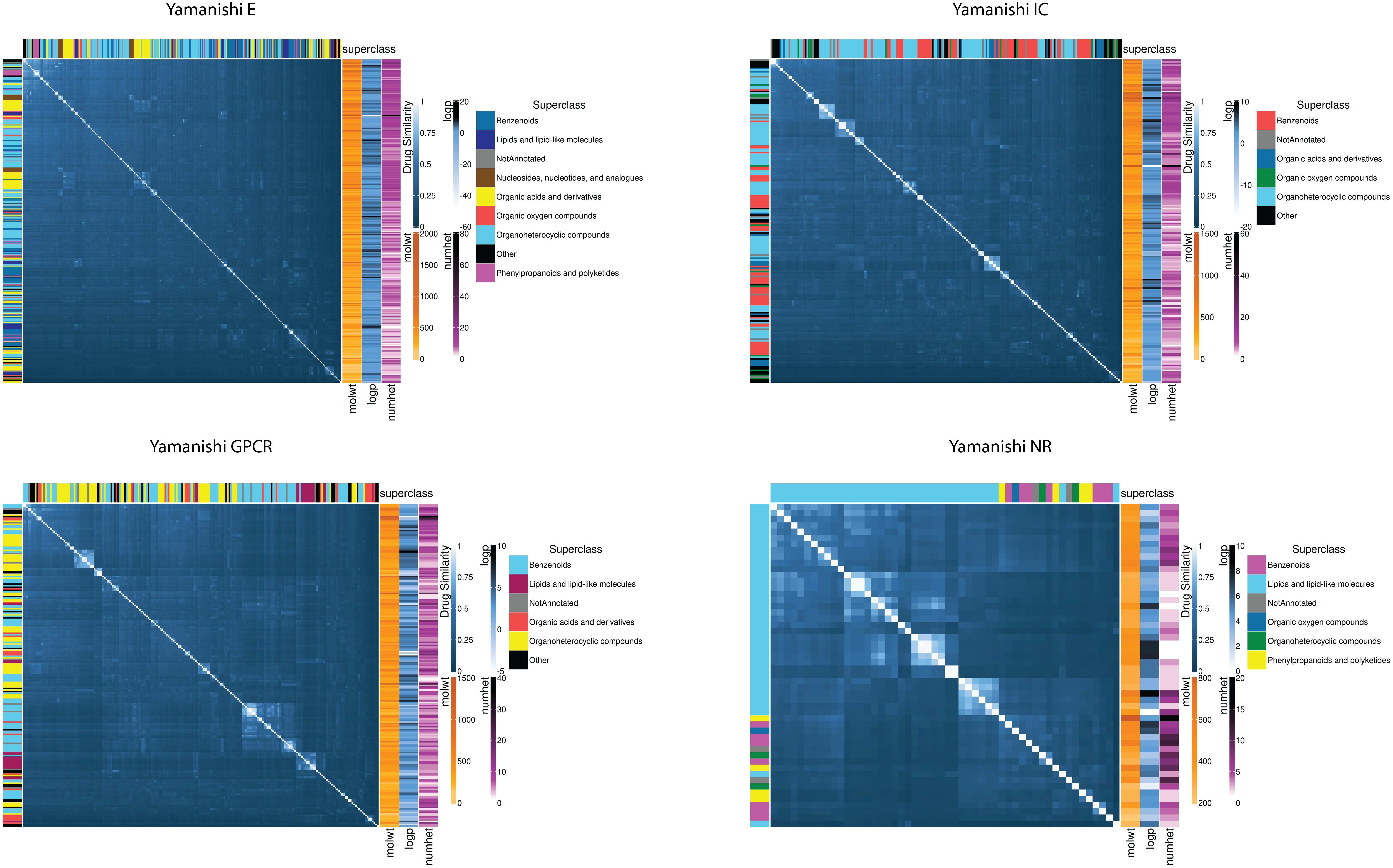}
\end{center}
\caption{Heatmaps of Tanimoto score for drugs of Yamanishi datasets calculated with euclidean distances. It is appreciated how some drugs form small clusters. The annotation represents the chemical classification of the drug (by superclass in Classyfire), and three different molecular descriptors the molecular weight (molwt), the logP, and the number of heteroatoms (numhet). }
\label{suppfig:s3_drugs_yamanishi_heatmap}
\end{figure}

\begin{figure}[H]
\begin{center}
\includegraphics[width=1\linewidth]{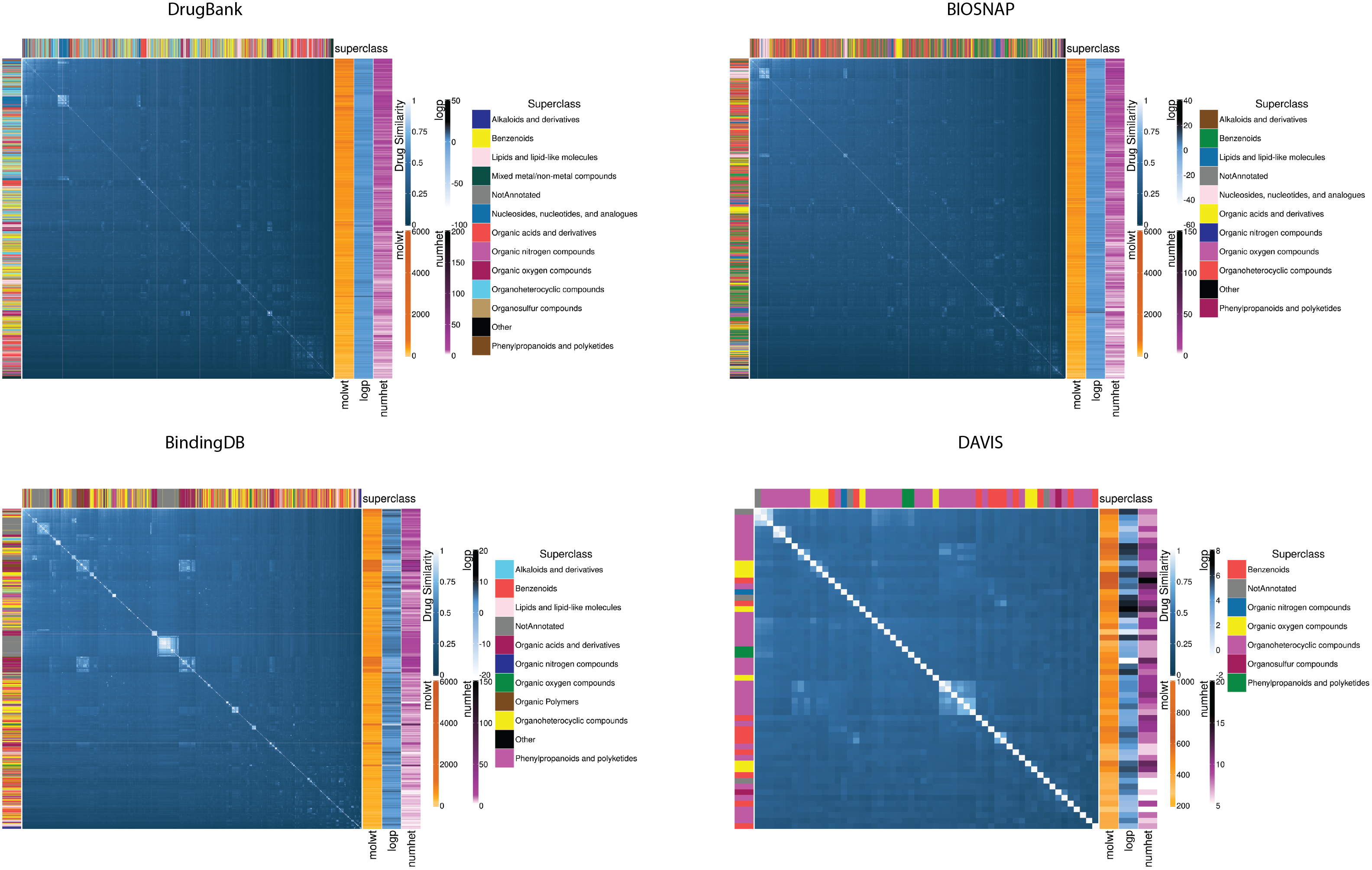}
\end{center}
\caption{Heatmaps of Tanimoto score for drugs of DrugBank, BIOSNAP, BindingDB and Davis datasets calculated with euclidean distances. It is appreciated how some drugs form small clusters. The annotation represents the chemical classification of the drug (by superclass in Classyfire), and three different molecular descriptors the molecular weight (molwt), the logP, and the number of heteroatoms (numhet).}
\label{suppfig:s3_drugs_dbbd_heatmap}
\end{figure}

\begin{figure}[H]
\begin{center}
\includegraphics[width=1\linewidth]{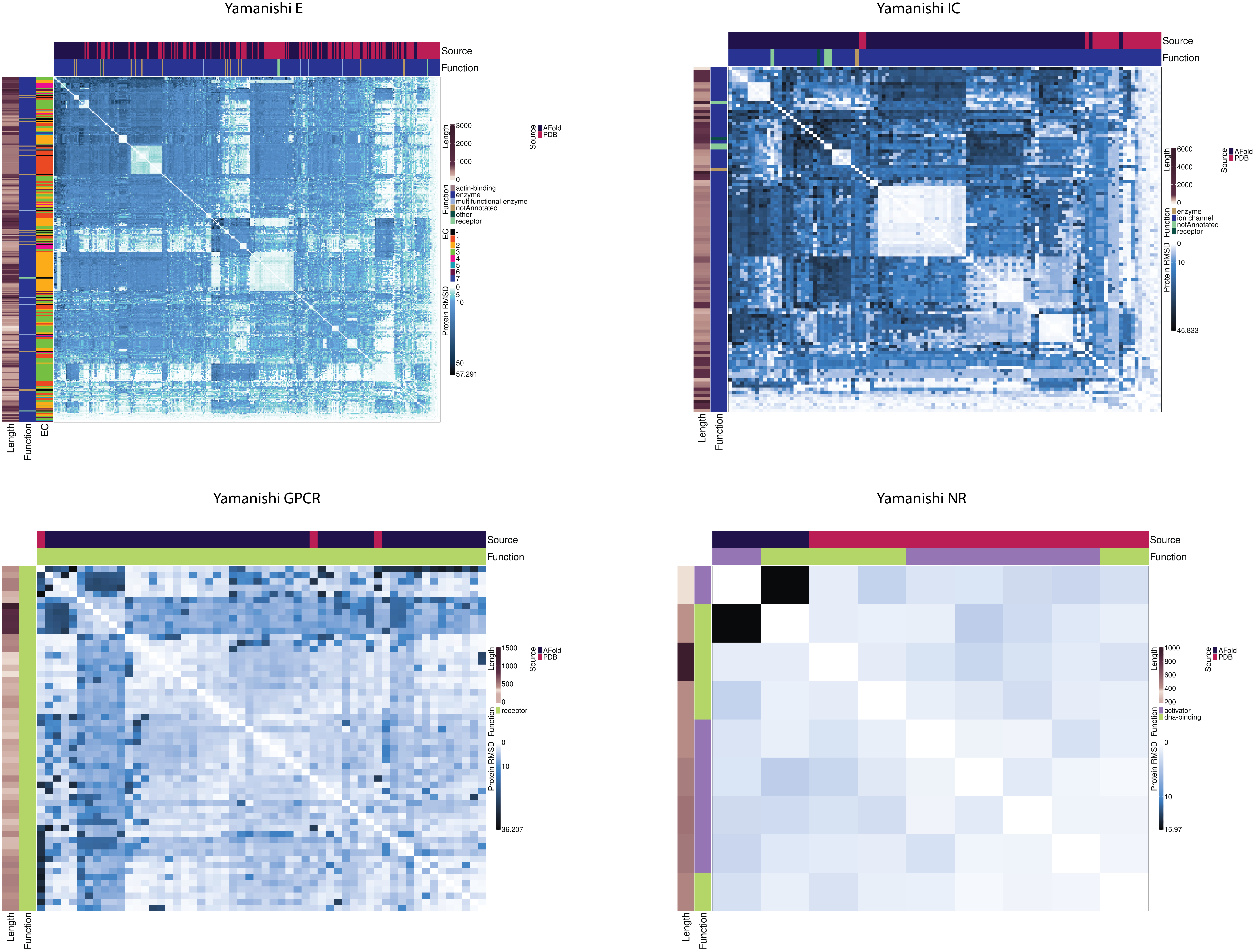}
\end{center}
\caption{Heatmaps of RMSD score for proteins of Yamanishi datasets calculated with Euclidean distances. The annotation represents the source of the protein, i.e., from where we downloaded the structure (PDB/AlphaFold), the molecular function of the protein, and the enzyme classification for the E dataset. Further, we added the sequence length in the left annotation in brown. In the heatmaps, proteins form several clusters. In E, this occurs especially for EC-1 (oxidoreductases) and EC-2 (transferases), further, smaller clusters of EC-3 (hydrolases) and EC-4 (lyases) also appear. Further, while in E and GPCR we found mixed the protein source, it does not happen for IC and NR.}
\label{suppfig:s4_proteins_yamanishi_heatmap}
\end{figure}

\begin{figure}[H]
\begin{center}
\includegraphics[width=1\linewidth]{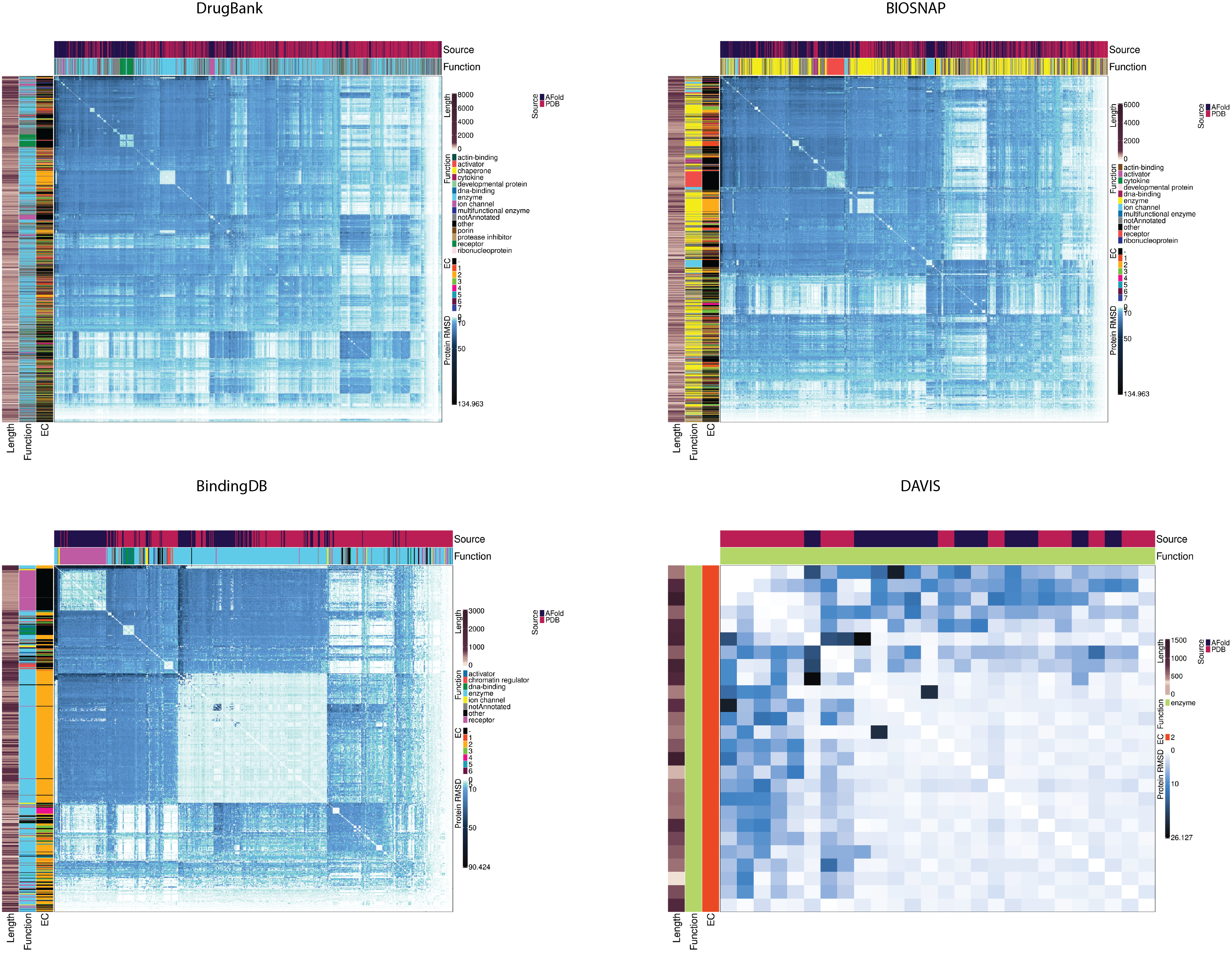}
\end{center}
\caption{Heatmaps of RMSD score for proteins of DrugBank, BIOSNAP, BindingDB and DAVIS datasets calculated with Euclidean distances. The annotation represents the source of the protein, i.e., from where we downloaded the structure (PDB/AlphaFold), the molecular function of the protein, and the enzyme classification for the E dataset. Further, we added the sequence length in the left annotation in brown. For the four datasets, we found a mixed distribution over clusterings of proteins independently of the source. Further, we shown how proteins clustering by protein family, which can be specially appreciated in BindingDB for receptors and enzymes, with brenda classification EC-2. }
\label{suppfig:s4_proteins_dbbd_heatmap}
\end{figure}


\begin{figure}[H]
\begin{center}
\includegraphics[width=0.5\linewidth]{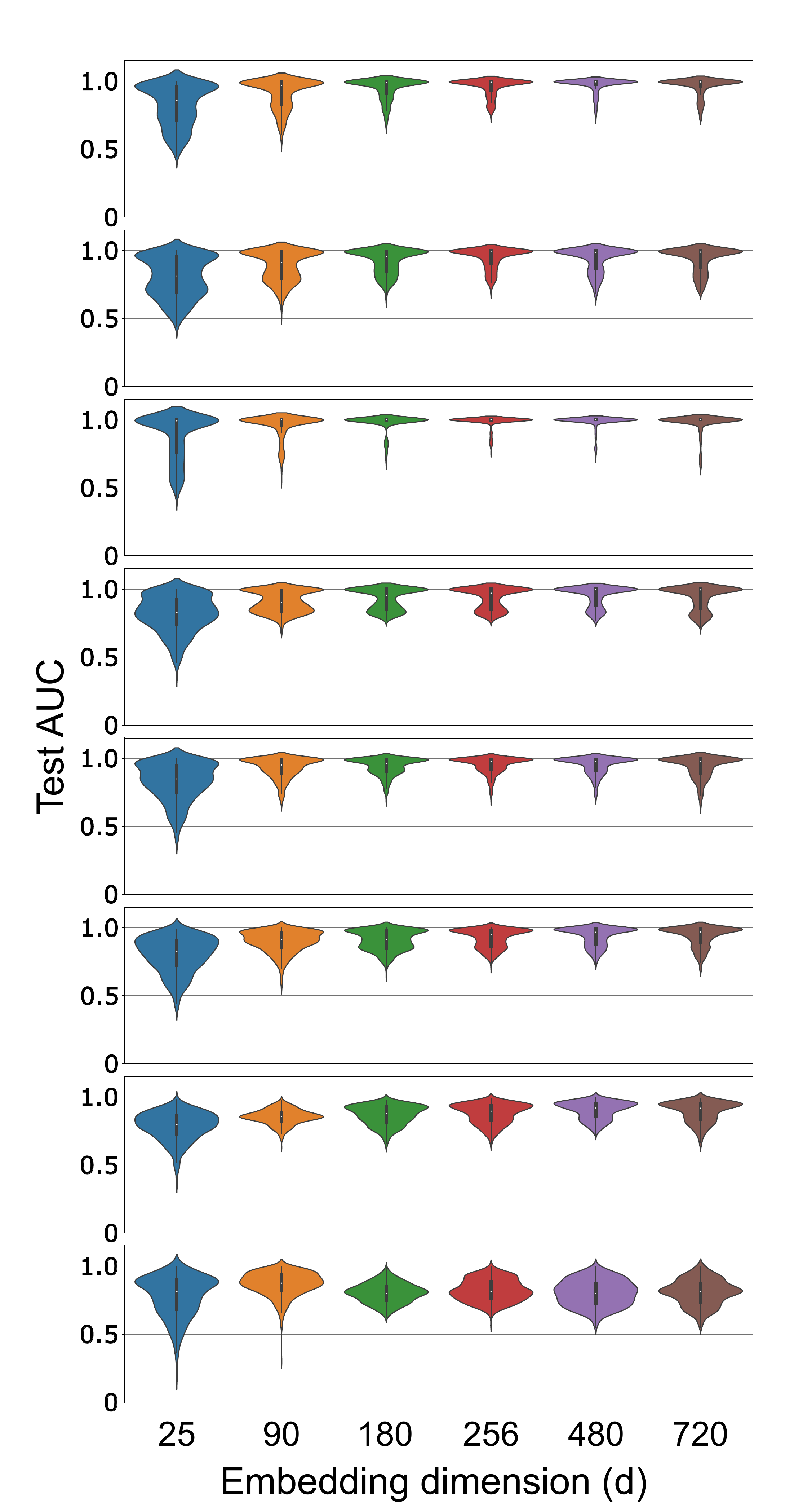}
\end{center}
\caption{Test AUC distribution across evaluated datasets, grouped by node2vec embedding dimension}
\label{suppfig:n2v_embdim}
\end{figure}

\begin{figure}[H]
\begin{center}
\includegraphics[width=0.8\linewidth]{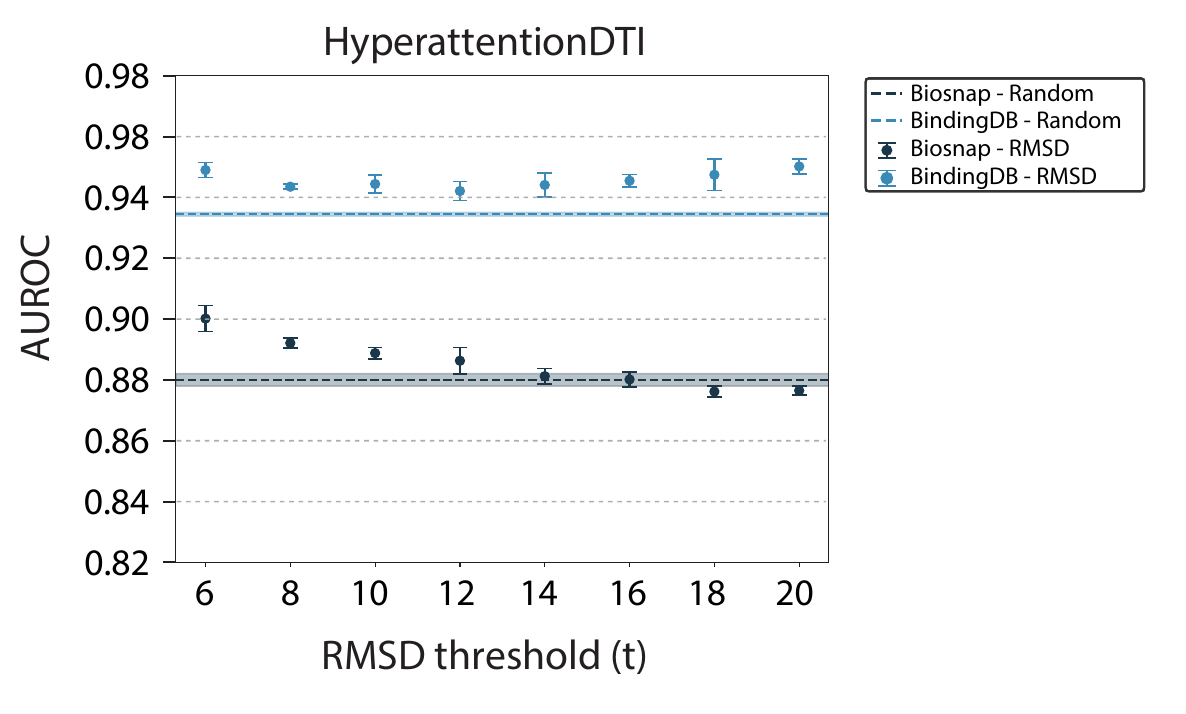}
\end{center}
\caption{AUROCs of the test split for the RMSD-based (threshold from 6 to 20 \AA) and random subsampling techniques when using HyperAttentionDTI model, on Biosnap and BindingDB datasets.}
\label{suppfig:rmsd_hyperattentiondti}
\end{figure}


\begin{figure}[H]
\begin{center}
\includegraphics[width=0.7\linewidth]{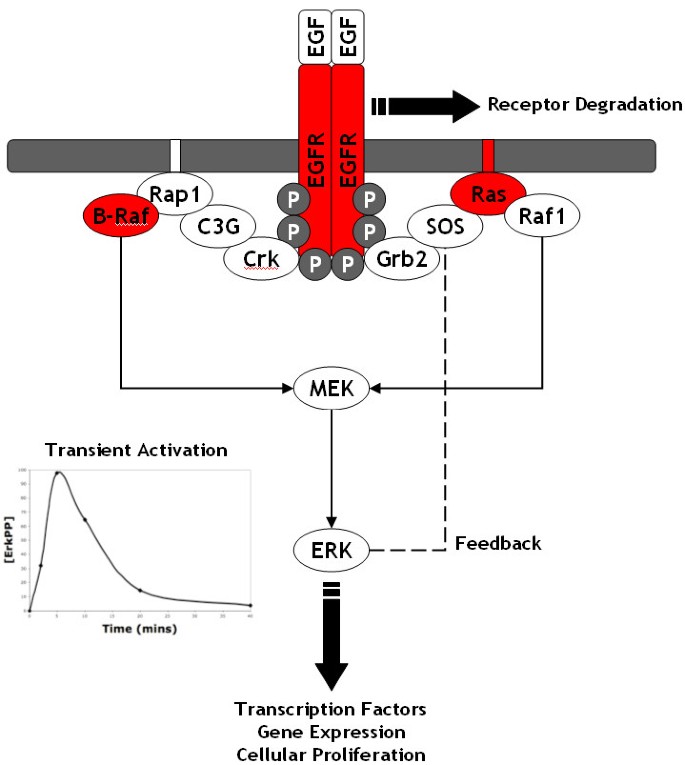}
\end{center}
\caption{EGF-ERK-pathway: This schematic illustrates the EGF-activated ERK pathway, starting with EGF binding to EGFR and concluding with ERK activation. Activated ERK has multiple targets in the cytoplasm and nucleus, including numerous transcription factors, therefore directly affecting gene expression and influence cellular growth. Source: https://doi.org/10.1186/1752-0509-3-100}
\label{suppfig:EGF-ERK-pathway}
\end{figure}

\begin{figure}[H]
\begin{center}
\includegraphics[width=1\linewidth]{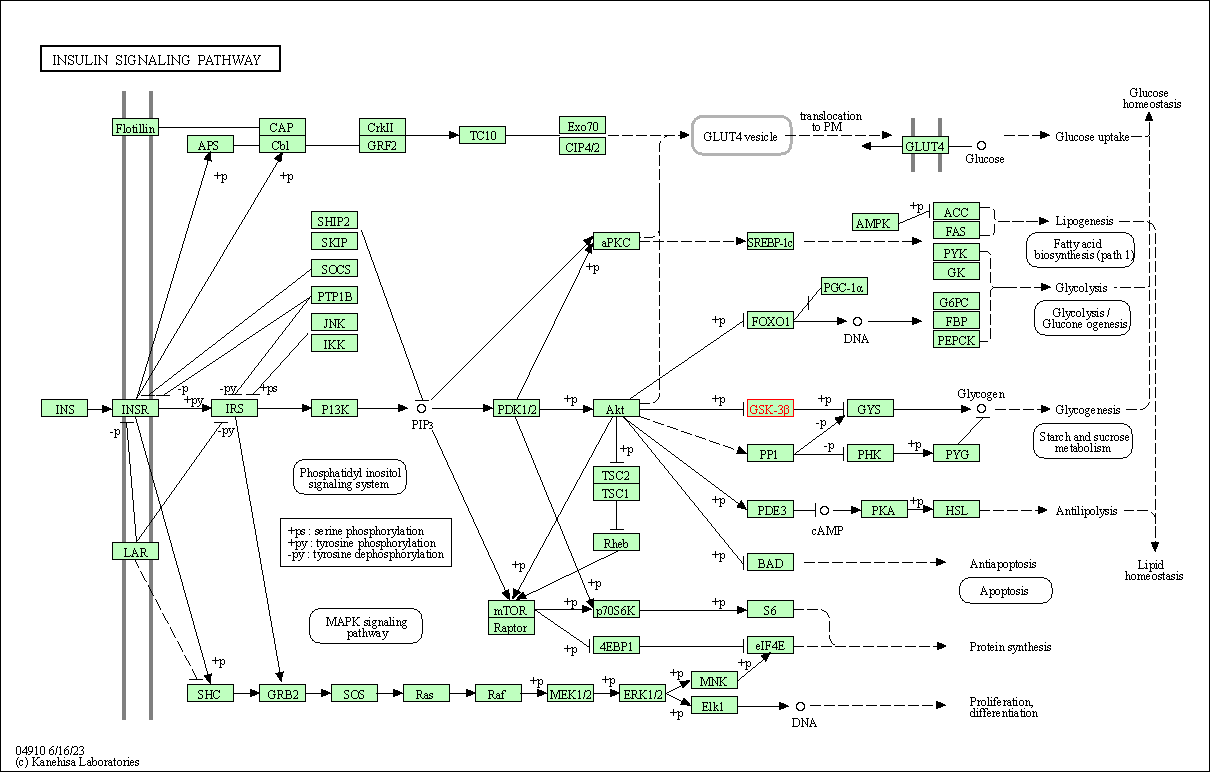}
\end{center}
\caption{Insulin signaling pathway: GSK3$\upbeta$, a multifaceted kinase, plays a critical role in regulating various cellular processes, including glycogen metabolism, cell growth, and apoptosis. S6, a vital component of the ribosome, is central to protein synthesis and cellular growth. By phosphorylating S6 into S6K, GSK3$\upbeta$ can modulate the activity of the ribosome and, consequently, protein synthesis. Source: KEGG signaling pathways \url{https://www.genome.jp/pathway/hsa04910+2932}.}
\label{suppfig:insulin-pathway}
\end{figure}


\clearpage
\section*{Appendix Tables}

\counterwithin{table}{section}
\setcounter{table}{0}

\renewcommand{\tablename}{Appendix Table}
\renewcommand{\thetable}{A\arabic{table}}

\begin{table}[ht]
\caption{\textbf{State-of-the-art DTI networks statistics}. The density of a graph represents the proportion of edges present in the graph to the total number of edges that could possibly exist in the graph. The number of connected components represents the count of isolated subgraphs within the network.}
\centering
\resizebox{\columnwidth}{!}{%

\begin{tabular}{lcccccccc}
 & \textbf{DrugBank} & \textbf{BIOSNAP}  & \textbf{BindingDB} & \textbf{DAVIS} & \begin{tabular}[c]{@{}c@{}}\textbf{Yamanishi}\\ \textbf{E}\end{tabular}  & \begin{tabular}[c]{@{}c@{}}\textbf{Yamanishi}\\ \textbf{IC}\end{tabular}  & \begin{tabular}[c]{@{}c@{}}\textbf{Yamanishi}\\ \textbf{GPCR}\end{tabular}  & \begin{tabular}[c]{@{}c@{}}\textbf{Yamanishi}\\ \textbf{NR}\end{tabular} 
\\ \hline \\
Number of drugs & 8042 & 5017 & 3085 & 65 & 445 & 210 & 223 & 54 \\
Number of proteins & 5141 & 2324 & 719 & 314 & 664 & 204 & 95 & 26 \\
Total number of nodes & 13183 & 7341 & 3804 & 379 & 1109 & 414 & 318 & 80 \\
Total number of edges & 27861 & 15138 & 5938 & 1048 & 2926 & 1476 & 635 & 90 \\
Density (\%) &  0.03 & 0.06 & 0.08 & 1.46 & 0.48 & 1.73 & 1.26 & 2.85\\
\# of connected components & 490 & 205 & 232 & 1 & 44 & 3 & 19 & 10\\
\end{tabular}}
\label{tab:statistics}
\end{table}

\begin{table}[htb]
\caption{\textbf{Comparison across DTI prediction methods}. Evaluation considers various criteria, such as the language used, the splitting or the utilized validation technique. Upper and lower groups illustrate transductive and inductive methods, respectively.}

\begin{center}
\resizebox{\columnwidth}{!}{%
\begin{tabular}{lllllllll}
&  & Complete & Issue-free & GitHub & Default & Data & Time & Validation \\
Method & Language & Code & Code & Medal & Splitting & Availability & Consumption & (type) \vspace{0.1cm} \\
\hline \\
DTiNet & Matlab & $\checkmark$ & $\checkmark$ & $\times$ & Unconstrained 10 folds CV & $\times$ & Very Fast & Experimental, In-sillico \\
DDR & Python & $\times$ & $\checkmark$ & \faTrophy{} & Sp, Sd, St & $\times$ & Slow & Bibliographic \\
DTi-GEMS & Python & $\times$ & $\times$ & $\times$ & Unconstrained 10 times TVT & $\times$ & Fast & In-sillico \\
DTi2Vec & Python & $\times$ & $\times$ & $\times$ & Unconstrained 10 times TVT & $\times$ & Slow & Bibliographic \vspace{0.15cm} \\ \hdashline \vspace{-0.05cm} \\
NeoDTI & Python & $\times$ & $\times$ & $\times$ & Unconstrained 10 times TVT & $\checkmark$ & Slow & Bibliographic \\
Moltrans & Python & $\times$ & $\checkmark$ & $\times$ & Unconstrained 5 times TVT & $\times$ & Fast & $\times$ \\
HyperAttentionDTI & Python & $\times$ & $\times$ & $\times$ & Sp, Sd, St, Sh & $\times$ & Fast & $\times$ \\
EEG-DTI & Python & $\times$ & $\times$ & $\times$ & Unconstrained 10 folds CV & $\checkmark$ & Fast & $\times$ \\
\end{tabular}
}
\begin{flushleft}
    \footnotesize{\tiny \textit{TVT}: train-validation-test, \textit{CV}: Cross-Validation}
    \end{flushleft}
\end{center}
\label{tab:model-comparison}
\end{table}

\begin{table}[htb]
\centering
\caption{Default Splits AUPRC benchmarking for the evaluated DTI prediction models.}
\resizebox{\columnwidth}{!}{%
\begin{tabular}{lcccccccc}
\toprule
\textbf{Method/Dataset} & DrugBank & BIOSNAP & BindingDB & Davis et al & E & IC & GPCR & NR \\
\midrule
DDR & OOT & OOT & OOT & $0.679 \pm 0.049$ & $0.911 \pm 0.010$ & $0.934 \pm 0.013$ & $0.818 \pm 0.036$ & $0.785 \pm 0.040$ \\
DTi2Vec & OOR & OOR & $0.920 \pm 0.005$ & $0.390 \pm 0.060$ & $0.970 \pm 0.012$ & $0.960 \pm 0.009$ & $0.830 \pm 0.058$ & $0.750 \pm 0.072$ \\
DTi-GEMS & OOR & OOR & $0.323 \pm 0.451$ & $0.386 \pm 0.210$ & $0.820 \pm 0.278$ & $0.873 \pm 0.092$ & $0.731 \pm 0.150$ & $0.626 \pm 0.170$ \\
DTINet & $0.857 \pm 0.001$ & $0.886 \pm 0.001$ & $0.873 \pm 0.006$ & $0.812 \pm 0.010$ & $0.918 \pm 0.004$ & $0.764 \pm 0.010$ & $0.803 \pm 0.006$ & $0.730 \pm 0.022$ \vspace{0.15cm} \\ \hdashline \vspace{-0.05cm} \\
NeoDTI & OOT & OOT & OOT & OOT & OOT & $0.917 \pm 0.005$ & $0.752 \pm 0.045$ & $0.450 \pm 0.145$ \\
Moltrans & $0.662 \pm 0.005$ & $0.645 \pm 0.004$ & $0.803 \pm 0.005$ & $0.530 \pm 0.039$ & $0.800 \pm 0.004$ & $0.772 \pm 0.007$ & $0.530 \pm 0.053$ & $0.421 \pm 0.018$ \\
HyperAttentionDTI & $0.776 \pm 0.046$ & $0.772 \pm 0.063$ & $0.910 \pm 0.051$ & $0.576 \pm 0.019$ & $0.922 \pm 0.062$ & $0.917 \pm 0.082$ & $0.648 \pm 0.057$ & $0.326 \pm 0.073$ \\
EEG-DTI & $0.880 \pm 0.007$ & $0.905 \pm 0.010$ & $0.890 \pm 0.030$ & $0.700 \pm 0.036$ & $0.955 \pm 0.009$ & $0.948 \pm 0.015$ & $0.882 \pm 0.047$ & $0.801 \pm 0.061$ \\

\bottomrule
\end{tabular}
}
\begin{flushleft}
    \footnotesize{\tiny OOT and OOR mean \textit{out of time} and \textit{out of RAM}, respectively.}
    \end{flushleft}
\label{suptab:auprc-default}
\end{table}


\begin{table}[ht]
  \centering
  \caption{Analyzed methodologies time consumption. }
  \resizebox{\columnwidth}{!}{%
    \begin{tabular}{lcccccccc}
    \hline
    \textbf{Model/Dataset}           & DrugBank-DTI   & BioSNAP        & BindingDB    & Davis et al   & E           & IC          & GPCR        & NR          \\ \hline
DDR                      & OOT           & OOT            & OOT          & 1h 33m 45s   & 5d 18h     & 3h 36m 21s  & 1h 16m 10s & 9m 17s    \\
DTi2Vec                  & OOR           & OOR            & 11h 50m     & 3h 17m 42s   & 3d 17h 18m & 2h 36m     & 1h 50m 49s & 1m 39s    \\
DTi-GEMS                 & OOR           & OOR            & 5h 25min    & 25m          & 22m     & 2min       & 1m 13s     & 20s       \\
DTINet                   & 39m 57s  & 21m 8s   & 3m 27s  & 1m 4s    & 6m 5s  & 2m 11s & 1m 11s & 8s    \vspace{0.15cm} \\ \hdashline \vspace{-0.05cm} \\ 
NeoDTI                   & OOT           & OOT            & OOT          & OOT           & OOT           & 1d 19h     & 1d 19h    & 1d 15h    \\
Moltrans                 & 6h 39m 18s   & 3h 36m 46s    & 1h 13m 47s  & 17m 49s      & 49m 14s    & 24m 13s    & 10m 42s   & 1m 34s    \\
HyperAttentionDTI             & 10h 10m 27s  & 5h 32m 50s    & 2h 31m 37s  & 31m 27s      & 1h 18m 29s & 45m 59s    & 18m 17s   & 2m 21s    \\
EEG-DTI    & 3d 13h 6m 45s & 3h 26m 57s  & 50m 21s     & 54m 39s       & 4h 30m 22s & 10m 16s & 328m 30s & 7m 47s\\ \hline
\end{tabular}
}
\begin{flushleft}
    \footnotesize{\tiny OOT and OOR mean \textit{out of time} and \textit{out of ram}, respectively.}
\end{flushleft}
\label{suptab:time-consumption} 
\end{table}


\begin{table}
    \centering
    \caption{Mean and standard deviation of the AUPRC values for the $S_p$ split, for each evaluated model and dataset.}
    \resizebox{\columnwidth}{!}{
    \begin{tabular}{lcccccccc}
        \toprule
         \textbf{Model/Dataset} & Yamanishi-NR & Yamanishi-IC & Yamanishi-GPCR & Yamanishi-E & DrugBank & DAVIS & BindingDB & BioSNAP \\
        \midrule
        DDR & $0.785 \pm 0.040$ & $0.934 \pm 0.013$ & $0.818 \pm 0.035$ & $0.910 \pm 0.010$ & OOT & $0.679 \pm 0.049$ & OOT & OOT \\
        DTI2Vec & NEET & $0.620 \pm 0.130$ & $0.760 \pm 0.128$ & $0.580 \pm 0.171$ & OOR & $0.58 \pm 0.098$ & $0.999 \pm 0.112$ & $0.57 \pm 0.132 $\\
        DTi-GEMS & $0.977 \pm 0.249$ & $0.953 \pm 0.254$ & $0.953 \pm 0.157$ & $0.996 \pm 0.128$ & OOR & OOR & OOR & OOR \\
        DTINet & $0.710 \pm 0.0133$ & $0.767 \pm 0.003$ & $0.767 \pm 0.009$ & $0.914 \pm 0.002$ & $0.830 \pm 0.001$ & $0.689 \pm 0.014$ & $0.800 \pm 0.002$ & $0.857 \pm 0.001$ \vspace{0.15cm} \\ \hdashline \vspace{-0.05cm} \\
        NeoDTI & $0.470 \pm 0.190$ & $0.893 \pm 0.018$ & $0.748 \pm 0.032$ & OOT & OOT & OOT & OOT & OOT \\
        Moltrans & $0.4206 \pm 0.014$ & $0.771 \pm 0.007$ & $0.530 \pm 0.053$ & $0.800 \pm 0.004$ & $0.620 \pm 0.005$ & $0.529 \pm 0.039$ & $0.803 \pm 0.005$ & $0.645 \pm 0.004$ \\
        HyperAttentionDTI & $0.326 \pm 0.073$ & $0.917 \pm 0.082$ & $0.648 \pm 0.057$ & $0.922 \pm 0.062$ & $0.776 \pm 0.046$ & $0.576 \pm 0.019$ & $0.910 \pm 0.051$ & $0.772 \pm 0.063$ \\
        EEG-DTI & $0.732 \pm 0.044$ & $0.962 \pm 0.007$ & $0.887 \pm 0.006$ & $0.962 \pm 0.002$ & $0.925 \pm 0.001$ & $0.747 \pm 0.015$ & $0.859 \pm 0.003$ & $0.928 \pm 0.002$ \\
        \bottomrule
    \end{tabular}
    }
    \begin{flushleft}
    \footnotesize{\tiny NEET, OOR and OOT stands for \textit{Not Enough Edges to Train}, \textit{Out of RAM} and \textit{Out of Time}, respectively.}
    \end{flushleft}
    \label{suptab:results_split_Sp}
\end{table}

\begin{table}
    \centering
    \caption{Mean and standard deviation of the AUPRC values for the $S_d$ split, for each evaluated model and dataset.}
    \resizebox{\columnwidth}{!}{
    \begin{tabular}{lcccccccc}
        \toprule
        \textbf{Model/Dataset} & Yamanishi-NR & Yamanishi-IC & Yamanishi-GPCR & Yamanishi-E & DrugBank & DAVIS & BindingDB & BioSNAP \\
        \midrule
        DDR & 0.642 $\pm$ 0.220 & 0.660 $\pm$ 0.095 & 0.608 $\pm$ 0.090 & 0.699 $\pm$ 0.650 & OOT & $0.462 \pm 0.014$ & OOT & OOT \\
        DTI2Vec & 0.73 $\pm$ 0.179 & 0.620 $\pm$ 0.012 & 0.785 $\pm$ 0.210 & 0.59 $\pm$ 0.132 & OOR & 0.630 $\pm$ 0.100 & OOR & 0.640 $\pm$ 0.140 \\
        DTi-GEMS & NEET & NEET & 0.796 $\pm$ 0.280 & 0.980 $\pm$ 0.120 & OOR & OOR & OOR & OOR \\
        DTINet & 0.635 $\pm$ 0.011 & 0.664 $\pm$ 0.005 & 0.741$\pm$ 0.011 & 0.844 $\pm$ 0.004 & 0.813 $\pm$ 0.001 & 0.556 $\pm$ 0.012 & 0.762 $\pm$ 0.008 & 0.841 $\pm$ 0.001 \vspace{0.15cm} \\ \hdashline \vspace{-0.05cm} \\
        NeoDTI & 0.118 $\pm$ 0.029 & 0.213 $\pm$ 0.060 & 0.271 $\pm$ 0.024 & OOT & OOT & OOT & OOT & OOT  \\
        Moltrans & 0.421 $\pm$ 0.038 & 0.437 $\pm$ 0.056 & 0.474 $\pm$ 0.071 & 0.411 $\pm$ 0.043 & 0.574 $\pm$ 0.025 & 0.333 $\pm$ 0.038 & 0.653 $\pm$ 0.044 & 0.530 $\pm$ 0.010 \\
        HyperAttentionDTI & 0.343 $\pm$ 0.020 & 0.638 $\pm$ 0.016 & 0.591 $\pm$ 0.036 & 0.611 $\pm$ 0.012 & 0.684 $\pm$ 0.005 & 0.395 $\pm$ 0.014 & 0.859 $\pm$ 0.007 & 0.680 $\pm$ 0.004 \\
        EEG-DTI & 0.747 $\pm$ 0.020 & 0.764 $\pm$ 0.017 & 0.832 $\pm$ 0.024 & 0.781 $\pm$ 0.020 & 0.892 $\pm$ 0.001 & 0.660 $\pm$ 0.033 & 0.779 $\pm$ 0.016 & 0.902 $\pm$ 0.002 \\
        \bottomrule
    \end{tabular}
    }
    \begin{flushleft}
    \footnotesize{\tiny NEET, OOR and OOT stands for \textit{Not Enough Edges to Train}, \textit{Out of RAM} and \textit{Out of Time}, respectively.}
    \end{flushleft}
    \label{suptab:results_split_Sd}
\end{table}

\begin{table}
    \centering
    \caption{Mean and standard deviation of the AUPRC values for the $S_t$ split, for each evaluated model and dataset.}
    \resizebox{\columnwidth}{!}{
    \begin{tabular}{lcccccccc}
        \toprule
        \textbf{Model/Dataset} & Yamanishi-NR & Yamanishi-IC & Yamanishi-GPCR & Yamanishi-E & DrugBank & DAVIS & BindingDB & BioSNAP \\
        \midrule
        DDR & 0.622 $\pm$ 0.300 & 0.786 $\pm$ 0.031 & 0.658 $\pm$ 0.103 & 795 $\pm$ 0.027 & OOT & 0.618 $\pm$ 0.44 & OOT & OOT \\
        DTI2Vec & 0.740 $\pm$ 0.117 & 0.580 $\pm$ 0.017 & 0.660 $\pm$ 0.170 & 0.690 $\pm$ 0.127 & OOR & 0.540 $\pm$ 0.130 & OOR & 0.440 $\pm$ 0.320 \\
        DTi-GEMS & NEET & NEET & 0.731 $\pm$ 0.297 & 0.976 $\pm$ 0.190 & OOR & OOR & OOR & OOR \\
        DTINet & 0.493 $\pm$ 0.017 & 0.856 $\pm$ 0.008 & 0.623 $\pm$ 0.006 & 0.860 $\pm$ 0.002 & 0.796 $\pm$ 0.001 & 0.649 $\pm$ 0.008 & 0.660 $\pm$ 0.010 & 0.815 $\pm$ 0.002 \vspace{0.15cm} \\ \hdashline \vspace{-0.05cm} \\
        NeoDTI & 0.111 $\pm$ 0.051 & 0.362 $\pm$ 0.12 & 0.163 $\pm$ 0.032 & OOT & OOT & OOT & OOT & OOT \\
        Moltrans & 0.266 $\pm$ 0.069 & 0.551 $\pm$ 0.064 & 0.488 $\pm$ 0.036 & 0.604 $\pm$ 0.038 & 0.660 $\pm$ 0.020 & 0.477 $\pm$ 0.018 & 0.561 $\pm$ 0.300 & 0.681 $\pm$ 0.049 \\
        HyperAttentionDTI & 0.294 $\pm$ 0.038 & 0.827 $\pm$ 0.012 & 0.565 $\pm$ 0.023 & 0.795 $\pm$ 0.022 & 0.775 $\pm$ 0.002 & 0.548 $\pm$ 0.023 & 0.617 $\pm$ 0.014 & 0.771 $\pm$ 0.002 \\
        EEG-DTI & 0.501 $\pm$ 0.048 & 0.690 $\pm$ 0.017 & 0.592 $\pm$ 0.011 & 0.604 $\pm$ 0.015 & 0.743 $\pm$ 0.001 & 0.512 $\pm$ 0.002 & 0.606 $\pm$ 0.012 & 0.721 $\pm$ 0.008 \\
        \bottomrule
    \end{tabular}}
    \begin{flushleft}
    \footnotesize{\tiny NEET, OOR and OOT stands for \textit{Not Enough Edges to Train}, \textit{Out of RAM} and \textit{Out of Time}, respectively.}
    \end{flushleft}
    \label{suptab:results_split_St}
\end{table}

\begin{table}[ht]
\centering
\caption{Grid-search parameters used for the baseline (N2V + NN) model. For the grid-search, they were used 6 different embedding space dimensions, 4 type of architectures, 4 different number of epochs, $BinaryCrossEntropy$ or $Focal Distance$ as the loss function, and $1/16$ and $1/64$ of the dataset size as the batch size.}
\begin{tabular}{ll}
\hline
Parameter & Value \\
\hline
Embedding dimension & d25, d90, d180, d256, d480, d720 \\
Architecture & Type 1, Type 2, Type 3, Type 4 \\
Epochs & 2, 5, 10, 50 \\
Loss Function & BCE, Focal \\
Batch Size & 1/16, 1/64 \\
\hline
\end{tabular}
\label{supptable:n2v_gridsearch}
\end{table}

\clearpage
\bibliographystyle{unsrt}
\bibliography{bibliography}

\begin{thebibliography}{10}

\bibitem{patching2014}
Simon~G. Patching.
\newblock Surface plasmon resonance spectroscopy for characterisation of membrane protein--ligand interactions and its potential for drug discovery.
\newblock {\em Biochimica et Biophysica Acta (BBA) - Biomembranes}, 1838(1, Part A):43--55, 2014.
\newblock Structural and biophysical characterisation of membrane protein-ligand binding.

\bibitem{shuker1996}
Suzanne~B. Shuker, Philip~J. Hajduk, Robert~P. Meadows, and Stephen~W. Fesik.
\newblock Discovering high-affinity ligands for proteins: Sar by nmr.
\newblock {\em Science}, 274(5292):1531--1534, 1996.

\bibitem{dimasi2016}
Joseph~A. DiMasi, Henry~G. Grabowski, and Ronald~W. Hansen.
\newblock Innovation in the pharmaceutical industry: New estimates of r\&d costs.
\newblock {\em Journal of Health Economics}, 47:20--33, 2016.

\bibitem{bronstein2021geometric}
Michael~M Bronstein, Joan Bruna, Taco Cohen, and Petar Veli{\v{c}}kovi{\'c}.
\newblock Geometric deep learning: Grids, groups, graphs, geodesics, and gauges.
\newblock {\em arXiv preprint arXiv:2104.13478}, 2021.

\bibitem{zheng2013}
Mingyue Zheng, Xian Liu, Yuan Xu, Honglin Li, Cheng Luo, and Hualiang Jiang.
\newblock Computational methods for drug design and discovery: focus on china.
\newblock {\em Trends Pharmacol Sci}, 34(10):549--559, Oct 2013.

\bibitem{lavecchia2016}
Antonio Lavecchia and Carmen Cerchia.
\newblock In silico methods to address polypharmacology: current status, applications and future perspectives.
\newblock {\em Drug Discov Today}, 21(2):288--298, Feb 2016.

\bibitem{olayan2018ddr}
Rawan~S Olayan, Haitham Ashoor, and Vladimir~B Bajic.
\newblock {DDR: efficient computational method to predict drug--target interactions using graph mining and machine learning approaches}.
\newblock {\em Bioinformatics}, 34(7):1164--1173, 11 2017.

\bibitem{peng2021eegdti}
Jiajie Peng, Yuxian Wang, Jiaojiao Guan, Jingyi Li, Ruijiang Han, Jianye Hao, Zhongyu Wei, and Xuequn Shang.
\newblock An end-to-end heterogeneous graph representation learning-based framework for drug--target interaction prediction.
\newblock {\em Briefings in Bioinformatics}, 22(5), 2021.

\bibitem{zhao2021hyperatt}
Qichang Zhao, Haochen Zhao, Kai Zheng, and Jianxin Wang.
\newblock {HyperAttentionDTI: improving drug--protein interaction prediction by sequence-based deep learning with attention mechanism}.
\newblock {\em Bioinformatics}, 38(3):655--662, 10 2021.

\bibitem{yamanishi2008prediction}
Yoshihiro Yamanishi, Michihiro Araki, Alex Gutteridge, Wataru Honda, and Minoru Kanehisa.
\newblock Prediction of drug--target interaction networks from the integration of chemical and genomic spaces.
\newblock {\em Bioinformatics}, 24(13):i232--i240, 2008.

\bibitem{wishart2006drugbank}
David~S Wishart, Craig Knox, An~Chi Guo, Savita Shrivastava, Murtaza Hassanali, Paul Stothard, Zhan Chang, and Jennifer Woolsey.
\newblock Drugbank: a comprehensive resource for in silico drug discovery and exploration.
\newblock {\em Nucleic acids research}, 34(suppl\_1):D668--D672, 2006.

\bibitem{zitnik2018biosnap}
Sagar~Maheshwari Marinka~Zitnik, Rok~Sosi\v{c} and Jure Leskovec.
\newblock {BioSNAP Datasets}: {Stanford} biomedical network dataset collection.
\newblock \url{http://snap.stanford.edu/biodata}, 2018.

\bibitem{davis2011comprehensive}
Mindy~I Davis, Jeremy~P Hunt, Sanna Herrgard, Pietro Ciceri, Lisa~M Wodicka, Gabriel Pallares, Michael Hocker, Daniel~K Treiber, and Patrick~P Zarrinkar.
\newblock Comprehensive analysis of kinase inhibitor selectivity.
\newblock {\em Nature biotechnology}, 29(11):1046--1051, 2011.

\bibitem{liu2007bindingdb}
Tiqing Liu, Yuhmei Lin, Xin Wen, Robert~N Jorissen, and Michael~K Gilson.
\newblock Bindingdb: a web-accessible database of experimentally determined protein--ligand binding affinities.
\newblock {\em Nucleic acids research}, 35(suppl\_1):D198--D201, 2007.

\bibitem{dunn2022}
Timothy~B. Dunn, Gustavo~M. Seabra, Taewon~David Kim, K.~Eur{\'\i}dice Ju{\'a}rez-Mercado, Chenglong Li, Jos{\'e}~L. Medina-Franco, and Ram{\'o}n~Alain Miranda-Quintana.
\newblock Diversity and chemical library networks of large data sets.
\newblock {\em Journal of Chemical Information and Modeling}, 62(9):2186--2201, 2022.

\bibitem{mencher2005}
Simon~K. Mencher and Long~G. Wang.
\newblock Promiscuous drugs compared to selective drugs (promiscuity can be a virtue).
\newblock {\em BMC Clinical Pharmacology}, 5(1):3, 2005.

\bibitem{wen2017deep}
Ming Wen, Zhimin Zhang, Shaoyu Niu, Haozhi Sha, Ruihan Yang, Yonghuan Yun, and Hongmei Lu.
\newblock Deep-learning-based drug--target interaction prediction.
\newblock {\em Journal of proteome research}, 16(4):1401--1409, 2017.

\bibitem{ozturk2018deepdta}
Hakime {\"O}zt{\"u}rk, Arzucan {\"O}zg{\"u}r, and Elif Ozkirimli.
\newblock Deepdta: deep drug--target binding affinity prediction.
\newblock {\em Bioinformatics}, 34(17):i821--i829, 2018.

\bibitem{luo2017dtinet}
Yunan Luo, Xinbin Zhao, Jingtian Zhou, Jinglin Yang, Yanqing Zhang, Wenhua Kuang, Jian Peng, Ligong Chen, and Jianyang Zeng.
\newblock A network integration approach for drug-target interaction prediction and computational drug repositioning from heterogeneous information.
\newblock {\em Nat Commun}, 8(1):573, 2017.

\bibitem{thafar2020dtigems}
Maha~A. Thafar, Rawan~S. Olayan, Haitham Ashoor, Somayah Albaradei, Vladimir~B. Bajic, Xin Gao, Takashi Gojobori, and Magbubah Essack.
\newblock Dtigems+: drug--target interaction prediction using graph embedding, graph mining, and similarity-based techniques.
\newblock {\em Journal of Cheminformatics}, 12(1):44, 2020.

\bibitem{thafar2021dti2vec}
Maha~A. Thafar, Rawan~S. Olayan, Somayah Albaradei, Vladimir~B. Bajic, Takashi Gojobori, Magbubah Essack, and Xin Gao.
\newblock Dti2vec: Drug--target interaction prediction using network embedding and ensemble learning.
\newblock {\em Journal of Cheminformatics}, 13(1):71, 2021.

\bibitem{wan2018neodti}
Fangping Wan, Lixiang Hong, An~Xiao, Tao Jiang, and Jianyang Zeng.
\newblock {NeoDTI: neural integration of neighbor information from a heterogeneous network for discovering new drug--target interactions}.
\newblock {\em Bioinformatics}, 35(1):104--111, 07 2018.

\bibitem{huang2020moltrans}
Kexin Huang, Cao Xiao, Lucas~M Glass, and Jimeng Sun.
\newblock {MolTrans: Molecular Interaction Transformer for drug--target interaction prediction}.
\newblock {\em Bioinformatics}, 37(6):830--836, 10 2020.

\bibitem{kuhn2016sider}
Michael Kuhn, Ivica Letunic, Lars~Juhl Jensen, and Peer Bork.
\newblock The sider database of drugs and side effects.
\newblock {\em Nucleic Acids Res}, 44, 2016.

\bibitem{davis2021ctd}
Allan~Peter Davis, Cynthia~J Grondin, Robin~J Johnson, Daniela Sciaky, Jolene Wiegers, Thomas~C Wiegers, and Carolyn~J Mattingly.
\newblock Comparative toxicogenomics database (ctd): update 2021.
\newblock {\em Nucleic Acids Research}, 49(D1):D1138--D1143, 2020.

\bibitem{park2012flaws}
Yungki Park and Edward~M Marcotte.
\newblock Flaws in evaluation schemes for pair-input computational predictions.
\newblock {\em Nature methods}, 9(12):1134--1136, 2012.

\bibitem{pahikkala2014}
Tapio Pahikkala, Antti Airola, Sami Pietil{\"a}, Sushil Shakyawar, Agnieszka Szwajda, Jing Tang, and Tero Aittokallio.
\newblock {Toward more realistic drug--target interaction predictions}.
\newblock {\em Briefings in Bioinformatics}, 16(2):325--337, 04 2014.

\bibitem{grover2016node2vec}
Aditya Grover and Jure Leskovec.
\newblock node2vec: Scalable feature learning for networks.
\newblock In {\em Proceedings of the 22nd ACM SIGKDD international conference on Knowledge discovery and data mining}, pages 855--864, 2016.

\bibitem{siltberg2011evolution}
Jessica Siltberg-Liberles, Johan~A Grahnen, and David~A Liberles.
\newblock The evolution of protein structures and structural ensembles under functional constraint.
\newblock {\em Genes}, 2(4):748--762, 2011.

\bibitem{EGFR_validation}
Richard~J Orton, Michiel~E Adriaens, Amelie Gormand, Oliver~E Sturm, Walter Kolch, and David~R Gilbert.
\newblock Computational modelling of cancerous mutations in the egfr/erk signalling pathway.
\newblock {\em BMC systems biology}, 3(1):1--17, 2009.

\bibitem{zhang2006s6k1}
Hui~H Zhang, Alex~I Lipovsky, Christian~C Dibble, Mustafa Sahin, and Brendan~D Manning.
\newblock S6k1 regulates gsk3 under conditions of mtor-dependent feedback inhibition of akt.
\newblock {\em Molecular cell}, 24(2):185--197, 2006.

\bibitem{miner}
Stanford-SNAP-Group.
\newblock Miner: Gigascale multimodal biological network.
\newblock {\em GitHub Repository}, 2017.

\bibitem{tdcdatabase}
Kexin Huang, Tianfan Fu, Wenhao Gao, Yue Zhao, Yusuf Roohani, Jure Leskovec, Connor~W. Coley, Cao Xiao, Jimeng Sun, and Marinka Zitnik.
\newblock Therapeutics data commons: Machine learning datasets and tasks for drug discovery and development, 2021.

\bibitem{zong2019}
Nansu Zong, Rachael Sze~Nga Wong, Yue Yu, Andrew Wen, Ming Huang, and Ning Li.
\newblock {Drug--target prediction utilizing heterogeneous bio-linked network embeddings}.
\newblock {\em Briefings in Bioinformatics}, 22(1):568--580, 12 2019.

\bibitem{zhang2022}
Junjun Zhang and Minzhu Xie.
\newblock Graph regularized non-negative matrix factorization with prior knowledge consistency constraint for drug--target interactions prediction.
\newblock {\em BMC Bioinformatics}, 23(1):564, 2022.

\bibitem{kanehisa2006kegg}
Minoru Kanehisa, Susumu Goto, Masahiro Hattori, Kiyoko~F Aoki-Kinoshita, Masumi Itoh, Shuichi Kawashima, Toshiaki Katayama, Michihiro Araki, and Mika Hirakawa.
\newblock From genomics to chemical genomics: new developments in kegg.
\newblock {\em Nucleic Acids Res}, 34(Database issue):D354--7, Jan 2006.

\bibitem{schomburg2004brenda}
Ida Schomburg, Antje Chang, Christian Ebeling, Marion Gremse, Christian Heldt, Gregor Huhn, and Dietmar Schomburg.
\newblock Brenda, the enzyme database: updates and major new developments.
\newblock {\em Nucleic Acids Res}, 32(Database issue):D431--3, Jan 2004.

\bibitem{gunther2008supertarget}
Stefan G{\"u}nther, Michael Kuhn, Mathias Dunkel, Monica Campillos, Christian Senger, Evangelia Petsalaki, Jessica Ahmed, Eduardo~Garcia Urdiales, Andreas Gewiess, Lars~Juhl Jensen, Reinhard Schneider, Roman Skoblo, Robert~B Russell, Philip~E Bourne, Peer Bork, and Robert Preissner.
\newblock Supertarget and matador: resources for exploring drug-target relationships.
\newblock {\em Nucleic Acids Res}, 36(Database issue):D919--22, Jan 2008.

\bibitem{faers}
US~Food and Drug Administration.
\newblock Questions and answers on fda's adverse event reporting system (faers).
\newblock {\em Washington: US Department of Health and Human Services}, 2018.

\bibitem{keshava2009hprd}
T~S Keshava~Prasad, Renu Goel, Kumaran Kandasamy, Shivakumar Keerthikumar, Sameer Kumar, Suresh Mathivanan, Deepthi Telikicherla, Rajesh Raju, Beema Shafreen, Abhilash Venugopal, Lavanya Balakrishnan, Arivusudar Marimuthu, Sutopa Banerjee, Devi~S Somanathan, Aimy Sebastian, Sandhya Rani, Somak Ray, C~J Harrys~Kishore, Sashi Kanth, Mukhtar Ahmed, Manoj~K Kashyap, Riaz Mohmood, Y~L Ramachandra, V~Krishna, B~Abdul Rahiman, Sujatha Mohan, Prathibha Ranganathan, Subhashri Ramabadran, Raghothama Chaerkady, and Akhilesh Pandey.
\newblock Human protein reference database--2009 update.
\newblock {\em Nucleic Acids Res}, 37, 2009.

\bibitem{kuhn2007stitch}
Michael Kuhn, Christian von Mering, Monica Campillos, Lars~Juhl Jensen, and Peer Bork.
\newblock Stitch: interaction networks of chemicals and proteins.
\newblock {\em Nucleic acids research}, 36(suppl\_1):D684--D688, 2007.

\bibitem{smedley2009biomart}
Damian Smedley, Syed Haider, Benoit Ballester, Richard Holland, Darin London, Gudmundur Thorisson, and Arek Kasprzyk.
\newblock Biomart--biological queries made easy.
\newblock {\em BMC genomics}, 10(1):1--12, 2009.

\bibitem{gaulton2012chembl}
Anna Gaulton, Louisa~J Bellis, A~Patricia Bento, Jon Chambers, Mark Davies, Anne Hersey, Yvonne Light, Shaun McGlinchey, David Michalovich, Bissan Al-Lazikani, et~al.
\newblock Chembl: a large-scale bioactivity database for drug discovery.
\newblock {\em Nucleic acids research}, 40(D1):D1100--D1107, 2012.

\bibitem{pytorch}
Adam Paszke, Sam Gross, Francisco Massa, Adam Lerer, James Bradbury, Gregory Chanan, Trevor Killeen, Zeming Lin, Natalia Gimelshein, Luca Antiga, Alban Desmaison, Andreas K\"{o}pf, Edward Yang, Zach DeVito, Martin Raison, Alykhan Tejani, Sasank Chilamkurthy, Benoit Steiner, Lu~Fang, Junjie Bai, and Soumith Chintala.
\newblock {\em PyTorch: An Imperative Style, High-Performance Deep Learning Library}.
\newblock Curran Associates Inc., Red Hook, NY, USA, 2019.

\bibitem{rdkit}
Greg Landrum, Paolo Tosco, Brian Kelley, Gedeck Sriniker, and Gedeck.
\newblock Rdkit: Open-source cheminformatics. version 2022.09.1.
\newblock 2022.

\bibitem{rcsbpdb}
Helen~M. Berman, John Westbrook, Zukang Feng, Gary Gilliland, T.~N. Bhat, Helge Weissig, Ilya~N. Shindyalov, and Philip~E. Bourne.
\newblock {The Protein Data Bank}.
\newblock {\em Nucleic Acids Research}, 28(1):235--242, 01 2000.

\bibitem{varadi2021afold}
Mihaly Varadi, Stephen Anyango, Mandar Deshpande, Sreenath Nair, Cindy Natassia, Galabina Yordanova, David Yuan, Oana Stroe, Gemma Wood, Agata Laydon, Augustin {\v Z}{\'\i}dek, Tim Green, Kathryn Tunyasuvunakool, Stig Petersen, John Jumper, Ellen Clancy, Richard Green, Ankur Vora, Mira Lutfi, Michael Figurnov, Andrew Cowie, Nicole Hobbs, Pushmeet Kohli, Gerard Kleywegt, Ewan Birney, Demis Hassabis, and Sameer Velankar.
\newblock {AlphaFold Protein Structure Database: massively expanding the structural coverage of protein-sequence space with high-accuracy models}.
\newblock {\em Nucleic Acids Research}, 50(D1):D439--D444, 11 2021.

\bibitem{jumper2021afold}
John Jumper, Richard Evans, Alexander Pritzel, Tim Green, Michael Figurnov, Olaf Ronneberger, Kathryn Tunyasuvunakool, Russ Bates, Augustin {\v Z}{\'\i}dek, Anna Potapenko, Alex Bridgland, Clemens Meyer, Simon A.~A. Kohl, Andrew~J. Ballard, Andrew Cowie, Bernardino Romera-Paredes, Stanislav Nikolov, Rishub Jain, Jonas Adler, Trevor Back, Stig Petersen, David Reiman, Ellen Clancy, Michal Zielinski, Martin Steinegger, Michalina Pacholska, Tamas Berghammer, Sebastian Bodenstein, David Silver, Oriol Vinyals, Andrew~W. Senior, Koray Kavukcuoglu, Pushmeet Kohli, and Demis Hassabis.
\newblock Highly accurate protein structure prediction with alphafold.
\newblock {\em Nature}, 596(7873):583--589, 2021.

\bibitem{PyMOL}
{Schr\"odinger, LLC}.
\newblock The {PyMOL} molecular graphics system, version~1.8.
\newblock November 2015.

\bibitem{vallejo2017}
Adrian Vallejo, Naiara Perurena, Elisabet Guruceaga, Pawel~K. Mazur, Susana Martinez-Canarias, Carolina Zandueta, Karmele Valencia, Andrea Arricibita, Dana Gwinn, Leanne~C. Sayles, Chen-Hua Chuang, Laura Guembe, Peter Bailey, David~K. Chang, Andrew Biankin, Mariano Ponz-Sarvise, Jesper~B. Andersen, Purvesh Khatri, Aline Bozec, E.~Alejandro Sweet-Cordero, Julien Sage, Fernando Lecanda, and Silve Vicent.
\newblock An integrative approach unveils fosl1 as an oncogene vulnerability in kras-driven lung and pancreatic cancer.
\newblock {\em Nature Communications}, 8(1):14294, 2017.

\bibitem{uniprot2023}
The~UniProt Consortium.
\newblock {UniProt: the Universal Protein Knowledgebase in 2023}.
\newblock {\em Nucleic Acids Research}, 51(D1):D523--D531, 11 2022.

\bibitem{classyfire}
Yannick Djoumbou~Feunang, Roman Eisner, Craig Knox, Leonid Chepelev, Janna Hastings, Gareth Owen, Eoin Fahy, Christoph Steinbeck, Shankar Subramanian, Evan Bolton, Russell Greiner, and David~S. Wishart.
\newblock Classyfire: automated chemical classification with a comprehensive, computable taxonomy.
\newblock {\em Journal of Cheminformatics}, 8(1):61, 2016.

\end{thebibliography}

\end{document}